%% file: main.tex
\documentclass{article}
\usepackage{iclr2027_conference,times} 
\iclrfinalcopy 

\usepackage[utf8]{inputenc}
\usepackage[T1]{fontenc}

\usepackage{url}
\usepackage{booktabs}
\usepackage{amsmath,amssymb}
\usepackage{amsthm}

\theoremstyle{remark}

\usepackage{graphicx}
\usepackage{multirow}
\usepackage{xcolor}
\usepackage[most]{tcolorbox}   
\graphicspath{{figs/}}
\newcommand{\dsname}{MoGround}   
\newcommand{\asmname}{\dsname-Retrieved}  
\newcommand{\dsbase}{\dsname-Base}       

\definecolor{myGreen}{RGB}{31, 157, 85}
\definecolor{myRed}{RGB}{207, 32, 39}

\definecolor{hypercolor}{rgb}{0.12, 0.47, 0.79}
\usepackage[pagebackref,breaklinks,colorlinks,allcolors=hypercolor]{hyperref}

\title{MoGround: Measuring and Mitigating Modality Distraction in Vision-Language Models}

\author{%
  \hfill Luca Zhou$^{1}$ \hfill Bo Zhao$^{2}$ \hfill Rose Yu$^{3}$ \hfill Emanuele Rodolà$^{1,4}$ \hfill Roberto Dessì$^{5}$ \hfill \\[1.5ex]
  {\small $^{1}$Sapienza University of Rome \quad $^{2}$Harvard University \quad $^{3}$UC San Diego}\\[0.5ex]
  {\small $^{4}$Paradigma \quad $^{5}$Not Diamond}
}

\begin{document}
\maketitle

\lhead{Preprint}
{\let\thefootnote\relax\footnotetext{Correspondence: \texttt{luca.zhou@uniroma1.it}. Code and data: \url{https://github.com/LuckerZOfficiaL/Modality-Distraction}.}}

\begin{abstract}
    We release \textbf{\dsname}, a vision-language dataset spanning four visual domains in which the answer to every question is guaranteed to be available from exactly one modality. This guarantee enables us to measure \textit{modality distraction}, the failure in which a model answers a question correctly from one modality alone and then flips to a wrong answer once irrelevant content from the other modality is added. Existing probes rarely establish single-modality answerability this way, making it hard to isolate distraction in the first place. Across seven open-source VLMs, we find that modality distraction is not universal but model-dependent. The weaker-grounded modality is the more distracted one ($r = +0.86$), and distraction scales inversely with grounding strength ($r = -0.90$). The single-modality guarantee also enables a mitigation method that needs to distinguish between relevant and irrelevant context. Trained on one split of \dsname{} alone, a weight-space robustness vector reduces distraction on all seven models by $29\%$ to $51\%$, at a cost of only $0.1$ average points of accuracy on standard multimodal tasks.

\end{abstract}

\section{Introduction}
\label{sec:intro}

\begin{figure}[t]
\centering
\includegraphics[width=\linewidth]{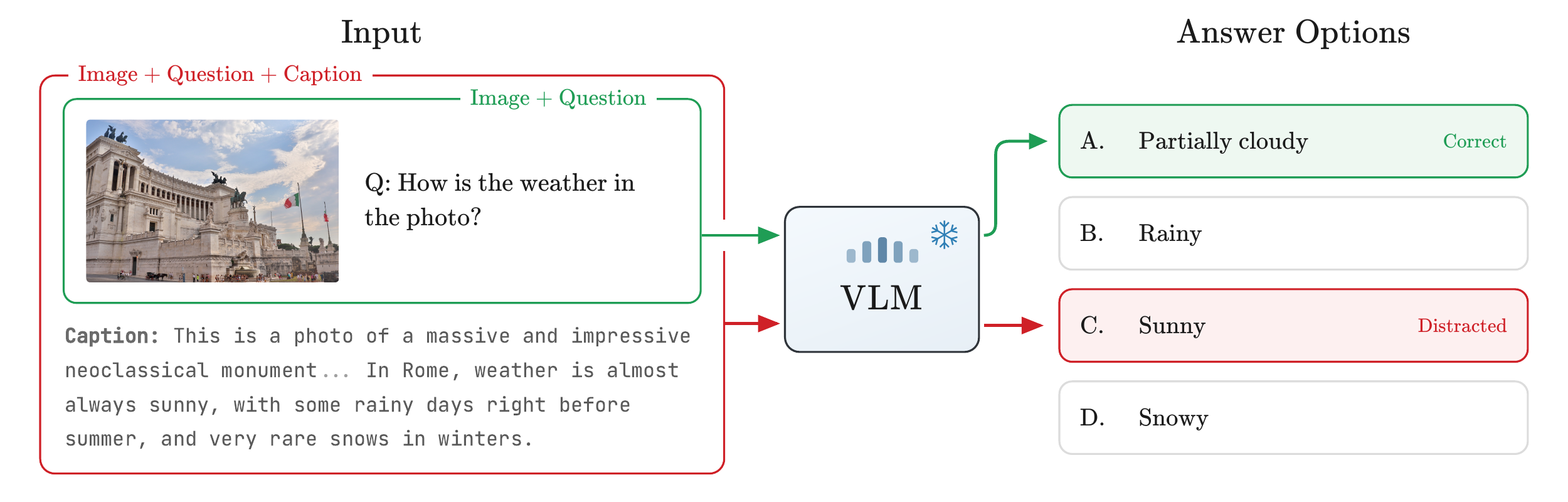}
\caption{
The same frozen VLM answers a question
twice. Given the image and the question alone, the model answers correctly (\textcolor{myGreen}{A, green}).
Once an irrelevant caption is added, it switches to a wrong answer (\textcolor{myRed}{C, red}), distracting the model from the correct option. The answer to the question can be found in the image without any explicit cue in the text, making the query vision-grounded.}
\label{fig:teaser}
\end{figure}

Vision-language models (VLMs) are widely reported to over-rely on language priors, answering from text even when the image
carries much information~\citep{lin2024revisiting,lee2024vlindbench,zhou-etal-2026-cats}.
We define modality \emph{distraction} as the phenomenon where a model
that answers a question correctly from one modality alone flips to a
wrong answer once irrelevant content from the other modality is added.
The literature is split on the direction of distraction. One line reports it as
\emph{asymmetric}, where irrelevant text degrades visual answering while irrelevant images
barely affect textual answering.
Another line treats it as \emph{bidirectional}~\citep{cai2025modality}. Neither side addresses the
question, because the direction has rarely been measured on data that isolates modality distraction.
The commonly used vision-language probes predominantly consist of photographs, and questions are
rarely validated to have the answer in only one modality. This conflates ``the model ignores the
image'' with ``the image was irrelevant''~\citep{chen2024mmstar,lin2024revisiting}. We find that
the direction is neither universal nor fixed, but a property of the individual model.
Figure~\ref{fig:teaser} illustrates an example of modality distraction in VLMs.

We start by creating the dataset required to isolate modality
distraction. We call it \dsname{}, for \emph{modality grounding}, since every
item is built so that its answer is grounded in one modality only. \dsbase{} is its main data pool,
a controllable probe set. Each item pairs an
image and a text description with a four-option multiple-choice question (MCQ), and an \emph{oracle filter} verifies that the item is answerable from exactly one modality: vision-grounded items from the image but not the text context, text-grounded items symmetrically. Items span four visually different domains: natural
photos, statistical charts, fine-art paintings, and medical radiology.
Two harder pools complete the dataset: a harder human-written set of questions for a set of images
(\dsname-Human) and an \asmname{} of image-question pairs drawn from established
benchmarks, one for visual question answering and one for reading comprehension. 
Our data creation pipeline allows to clearly quantify the phenomenon of distraction in VLMs, isolating items that a model solves from the grounded modality alone. 

Relying on \dsbase{}, we first quantify distraction on
seven commonly used open VLMs: every is affected by distraction, and non-natural domains such as charts and radiology show a larger performance drop. 
We then show that distraction is decided by a simple quantity, the model's grounding strength in the modality containing the answer, i.e.\ how accurately it answers from that modality alone. 
Finally, we show that a simple LoRA fine-tuning strategy on \dsname{} removes a large fraction of distraction at negligible capability cost.

Our contributions are the following:

\begin{itemize}

 \item We introduce \textbf{\dsbase{}, a dataset that isolates distraction.} To our
 knowledge, the first dataset in which every item is verified as
 answerable from exactly one modality. Verification is done by LLM
 judges and confirmed by a human audit. The dataset spans
 four visually different domains and ships with two harder subsets, one
 of them human-authored.

  \item The dataset supports a \textbf{simple remedy}. Training on \dsbase{}'s train split alone yields a robustness vector that reduces distraction on all seven backbones, transfers to pools it never saw, and comes at almost no general capability loss.
 
 \item Our \textbf{quantitative analysis shows that which modality is more
distractible varies across models}, and that it is predictable: the
weaker-grounded modality is the more distracted one. Underlying this,
distraction scales inversely with grounding strength,
consistent with irrelevant context eroding the model's confidence in an
answer it already has.

\end{itemize}

\section{Related Work}
\label{sec:related}
\paragraph{Modality bias in VLMs.} Vision-language models (VLMs) are
widely reported to over-rely on language
priors~\citep{lin2024revisiting,lee2024vlindbench}. The bias is not
limited to natural images: even when describing statistical plots,
models lean on the textual context and under-use the visual
signal~\citep{zhou-etal-2026-cats}. \citet{lee2024vlindbench} further observe that the bias decreases with
backbone scale; in our measurements, scale by itself does not predict
distraction (Appendix~\ref{app:horserace}), which distinguishes
distraction from the bias they measure.


\paragraph{Mitigating modality bias.} Existing benchmarks probe modality
bias in three main ways: with counterfactual or prior-violating
images~\citep{luo2025vilp,lee2024vlindbench}, by curating
``vision-indispensable'' items~\citep{chen2024mmstar}, or by creating
artificial conflicts between the two modalities, either to test which
one the model trusts~\citep{zhang2025modalitypref,ortu2025seeing} or to
mislead it with text that contradicts the
image~\citep{cai2025modality}. None of them verifies single modality grounding per
item. Curation-based sets assume visual dependency rather than test
it~\citep{chen2024mmstar,luo2025vilp} and, unlike our work, distraction sets apply no filter at all~\citep{yang2026defying,cai2025modality}.
Conflict-based probes also measure a
different quantity than we do: modality \emph{preference}, i.e., which
modality the model follows when the two disagree. Indeed,
\citet{cai2025modality} measure very different effects for unrelated;
\dsbase{} deliberately engineers no conflict. 
Three works come close to ours. \citet{shin2026surgcheck} also test
models with the image alone, the text alone, and both together, but
they use this protocol to audit existing surgical-VQA benchmarks in
that single domain; we use it to build a new multi-domain dataset in
which every item is verified to be answerable from one modality only.
\citet{singla2026seeorguess} check whether a question needs the image
simply by removing the image and seeing if accuracy drops, a weaker
check that does not verify what the text alone can answer.
\citet{sun2026distractbench} study the mirror setting of ours: they add
irrelevant \emph{visual} clutter to the image, whereas we add irrelevant
text. To our knowledge, no prior work isolates modality distraction
with an empirical study across models (seven backbones) and four
visual domains, together with a mechanistic analysis and a proposed intervention.

\paragraph{Modality interference.} \citet{cai2025modality} are the
closest work on the mitigation side: they fine-tune the interference
away in both directions, but treat it as a universal property of VLMs;
we show instead that it is model-dependent and explain the mechanism
behind it. \citet{yang2026defying} are the closest by failure mode:
their IR-VQA adds plausible but irrelevant input in both modalities and
mitigates it with a router that decides when to trust the added
context. However, their items are not verified to have the answer in only one
modality, so an accuracy drop cannot be attributed to distraction
alone; moreover, in our experiments a router of this kind only works
when it is told in advance which items are distracted.
\citet{hua2026tokenswap} train models to give the same answer when the
same content is presented as text or as image; we study the opposite
case, where the answer lives in exactly one modality and the model
abandons it. \citet{zhang2025modalitypref} show that a model's
\emph{global} preference for one modality can be steered with a single
activation direction, and \citet{huang2026amps} adapt that steering per
input. We find the complementary negative result: the per-item distraction
event is not controlled by any shared direction in our tests: the only
edit that repairs it uses the item's own clean activations, which no
deployed system has (Appendix~\ref{app:oneitem}). Finally, \citet{favero2024m3id} propose a
training-free alternative at decoding time: they run the model with and
without the image and push the answer toward what the image
contributes. The rule was designed against hallucination in long
captions, and we evaluate it as our strongest deployable baseline
(Appendix~\ref{app:oneitem}). 

\section{Data curation pipeline}
\label{sec:dataset}
We describe the creation pipeline for all three data splits we release, with concrete examples in Appendix~\ref{app:examples}. Table~\ref{tab:composition} gives the composition of all three data pools we release and describe next.    

\begin{figure}[t]
\begin{tcolorbox}[colback=cyan!5,colframe=cyan!40,arc=1mm,boxrule=0.3pt,width=\textwidth,top=3pt,bottom=3pt,left=5pt,right=5pt]
\noindent\begin{minipage}[t]{0.30\linewidth}\centering\vspace{0pt}\includegraphics[width=\linewidth,height=3.2cm,keepaspectratio]{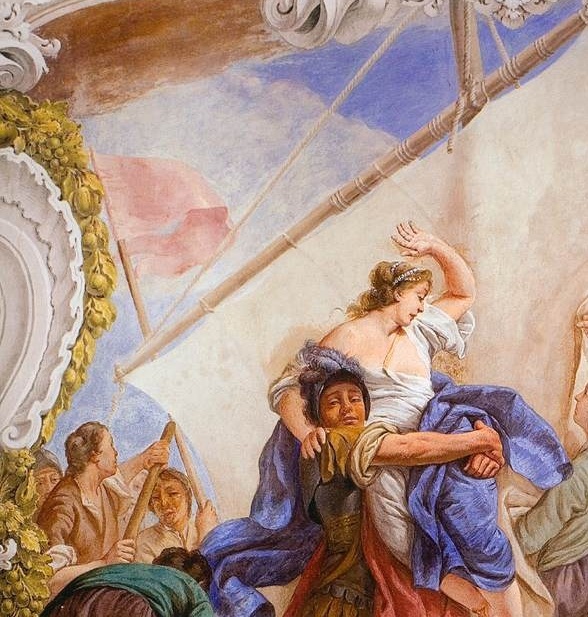}\end{minipage}\hfill
\begin{minipage}[t]{0.66\linewidth}\small\vspace{0pt}
\textbf{Caption.}~The two large scenes that extend between the doors on the walls in the ballroom are taken from Greek mythology and depict The Sacrifice of Iphigenia and The Abduction of Helen. [\dots] It depicts the abduction of Queen Helen [\dots]\\[3pt]\rule{\linewidth}{0.15pt}\\[3pt]
\textbf{Question.}~What color is the flag flying from the mast?\\[3pt]\rule{\linewidth}{0.15pt}\\[3pt]
\textbf{Options.}~A. White \quad \textbf{B. Pink~\checkmark} \quad C. Yellow \quad D. Blue
\end{minipage}
\end{tcolorbox}
\caption{A vision-grounded MoGround item from \textit{SemArt}, answerable via image but not via caption. The color of the flag is visible in the image but nowhere described in the caption.}
\label{fig:example} 
\end{figure}

\textbf{\dsbase{}.} We build \dsbase{} with a two-stage pipeline:
generate, then verify. Each candidate question starts from a real
image-caption pair. Each entry pairs an image, a caption, a question and four options, and we
call it an \emph{item}. In the first stage, an oracle LLM generates a
multiple-choice question whose answer lies in one chosen modality only.
For \textit{text-grounded} items, the oracle injects the answer into the
caption; for \textit{vision-grounded} items, the answer must be visible
in the image but not inferable from the caption. The question itself is always text. What verification decides is where the answer-bearing
evidence sits, in the image or in the context text, and we keep only items where it
sits in exactly one of the two. The second stage checks that this single-modality property
actually holds. A different oracle
answers each candidate under three input conditions: image-only ($V$),
text-only ($T$), and image+text ($VT$). We keep an item as
\emph{vision-grounded} iff $V$ and $VT$ are correct while $T$ fails, and
as \emph{text-grounded} under the symmetric rule. This behavioral filter
is what verifies single-modality grounding. The full pipeline is detailed in Appendix~\ref{app:pipeline}, with both oracle prompts in Appendix~\ref{app:oracle_prompts};
Figure~\ref{fig:example} shows an example question.

 We use Gemini 3 Flash and Claude Sonnet 4.6 as oracles and the
source image-caption pairs come from four datasets chosen to be visually
different: DCI~\citep{urbanek2023dci} (natural photos),
VisText~\citep{tang2023vistext} (statistical charts),
SemArt~\citep{garcia2018semart} (fine-art paintings), and
ROCO~\citep{ruckert2024rocov2} (medical radiology). 
Each source dataset is generated and verified by one oracle. Sonnet for DCI and VisText, Gemini for SemArt and ROCO.
The oracle filtering drops $74.5\%$ of the generated candidates, from $46.9\%$
on radiology to $85.9\%$ on charts. 


Three checks show that the single-modality property holds at the dataset level, not just in
one oracle's judgment. First, a cross-oracle check
ensures the filter is not an artifact of one particular oracle. We take
a subset of $240$ items ($60$ per source) and re-answer each
under the same three conditions, using the oracle that did \emph{not}
filter that item, so no oracle re-judges its own decision. The second
oracle agrees with the first on $94\%$ of items
(Appendix~\ref{app:threecond}). Second, oracle agreement is still machine
agreement, so two human annotators hand-check a random sample of
$250$ items, half vision- and half text-grounded. The two annotators agree on $98.4\%$ of the sample
(Appendix~\ref{app:audit_agree_sec}), and the single-modality property
holds for $97.6\%$ and $96.8\%$ of them, and at most $1.6\%$ leak the
answer into the wrong modality (Appendix~\ref{app:audit_base}). Third,
we check the benchmarked models themselves: across the seven models that we tried,
vision-grounded items are answered from the caption alone at
chance level, while the candidates the filter rejected score $12.7\%$
above chance.

\textbf{\dsname-Human: a human-written hard subset.} We also release
\textit{\dsname-Human}, $125$ captions written entirely by a human annotator on
for images from the same datasets above but on different instances than those used for \dsbase{}. The captions written to \emph{tempt} a wrong option. For instance, they may mention words
from the distractor options while describing unrelated objects or
relations, but a reader with only the caption must remain unable to
answer. 
We note that \textit{\dsname-Human is never used for selection or training}; it
serves only as a held-out test set for evaluation purposes.
Appendix~\ref{app:ex_human} provides four \dsname-Human samples.

\textbf{MoGround-Retrieved.} To add harder questions, we assemble another set from two
established benchmarks:
A-OKVQA~\citep{schwenk2022aokvqa} (vision-grounded VQA) and
RACE-high~\citep{lai2017race} (text-grounded reading comprehension), since difficulty was never a curation criterion for \dsbase{}.
Each item keeps its native grounding source (an image for vision items,
a passage for text items), and we pair it with an irrelevant distractor
from the other modality (a caption for vision items, an image for text
items). These questions are harder than \dsbase's because they were never written to be verified this way. A-OKVQA requires world knowledge about the depicted scene and RACE-High is exam-grade
reading comprehension, whereas \dsbase{} keeps an item only when the grounded modality already
answers it. Single-modality accuracy on this pool
spans $65\%$ to $97\%$ across backbones, against a \dsbase{} text side that sits near ceiling. Distractors are retrieved from CC3M~\citep{sharma2018conceptual}
by text similarity: each CC3M image is represented by its caption, and
captions are matched to the item's text with a sentence encoder
(\texttt{all-mpnet-base-v2}, \citealp{reimers2019sentencebert}).
Balanced subsampling yields $2{,}500$ vision $+$ $2{,}500$ text
$=5{,}000$ four-option questions. Retrieval alone cannot
\emph{guarantee} that a distractor is irrelevant, so we audit every
retrieved caption with Claude Sonnet 4.6 as the judge, in two independent passes that agree on
$95\%$ of a $120$-item overlap, and drop the $9.6\%$ that leak or
contradict an answer. After removing duplicate items, the final pool contains $4{,}757$ items
(protocol in Appendix~\ref{app:audit_retr}). Two independent human audits of $250$
pool items found one leak among the items the judge
kept, against leak rates of $36\%$ and $22\%$ among the ones it dropped
(Appendix~\ref{app:audit_retr}). We release \asmname{} alongside \dsbase{}, and every asymmetry
and transfer result in this paper is measured on it. Appendix~\ref{app:ex_assembled} shows two examples of \asmname{}.

\section{Measuring Modality Distraction}
\label{sec:leaderboard}

\dsbase{} is partitioned into a fixed 60/20/20 train/val/test split, so it can be
used both for evaluation and for training. In this section and in the
analyses of Sections~\ref{sec:asymmetry} and~\ref{sec:mechanism} we use
the combination of train, validation, and test splits since no training is involved. The train split is used only for the mitigation method
of Section~\ref{sec:mitigation}, where we use the validation set for parameter tuning and the test set for evaluation.

We evaluate seven open VLMs:
Qwen2-VL-2B~\citep{wang2024qwen2vl},
Qwen2.5-VL-3B/7B~\citep{bai2025qwen25vl},
InternVL3-8B~\citep{zhu2025internvl3},
LLaVA-1.5-7B~\citep{liu2024improvedllava},
LLaVA-NeXT-8B~\citep{liu2024llavanext}, and
LLaVA-OneVision-7B~\citep{li2024llavaonevision}.
The seven backbones differ in many ways, see architectural details in Appendix~\ref{app:backbones}.
Evaluation is done under the same three input conditions used in the data creation pipeline
($V$, $T$, $VT$). An answer is the option letter with the highest probability among A, B, C
and D. We measure \textbf{distraction} as a conditional flip rate. For instance, given a vision-grounded question that a model answers correctly
from the image alone, the datapoint is \emph{v-distracted} if the model
answers it wrongly once the irrelevant caption is added.
\emph{T-distraction} is symmetric.
Importantly, the conditioning is
\emph{model-relative}: each backbone distraction rate is computed only on the items that we assess it is able to answer from the grounded modality alone. Distraction therefore always measures a capability the model has and loses once irrelevant
context is added. A caption also repairs some items it could not answer from the image alone, so
the conditional rate is not an accuracy drop; Appendix~\ref{app:flips_sec} separates the two
directions. 
Results on \dsbase{} are reported in Table~\ref{tab:leaderboard} with confidence intervals in Appendix~\ref{app:cellcounts} (Table~\ref{tab:cell_cis}).

\begin{table}[t]
    \centering\scriptsize

    \label{tab:leaderboard}
\input{tables/leaderboard}
    \caption{Distraction results on \dsname{}-Base. Accuracy is computed with only one modality as input together with the question. V-distraction rate is the fraction of vision-grounded items answered correctly from the image alone but wrongly once a caption is added. Domain breakdown for T-grounded is in
Table~\ref{tab:tdistr_base}). Non:Nat Ratio is all non-photos domains vs photos v-distraction ratio.}
\end{table}

\textbf{Vision-grounded results.} Weaker backbones are the most distractible, and
non-natural domains (charts, paintings, radiology) are more distraction-prone than photos.
First, v-distraction rises from $3.4\%$ on InternVL3-8B to
$13.6\%$ on LLaVA-1.5-7B as V-only accuracy falls from $88\%$ to $46\%$.
Second, photos are the least distractible
domain in three of the seven models, and their v-distraction sits below
the mean of the three non-natural domains in five of seven. Charts and
radiology are the worst, averaging $11.3\%$ and $10.5\%$ across backbones
against $6.8\%$ on photos (per-domain breakdown in Appendix~\ref{app:perdomain}). These rates measure the effect of the
caption's \emph{content}, not of a longer prompt: when we replace the
true caption with text of the same length (another item's caption,
the same words shuffled, or content-free filler), the distraction rate
always drops (Appendix~\ref{app:placebo}).

\textbf{Text-grounded results.} T-distraction is ${\approx}0$ for
every backbone: an added irrelevant image almost never flips a
text-grounded answer. This reflects how the data was built rather than
a property of the models. Our pipeline verifies that the answer lies in
the text, but it does not control for how \emph{difficult} the question
is. The captions turn out to be too revealing for the questions: every model answers
text-grounded items with above $99\%$ accuracy, and we find that those answers are too far from the decision boundary for an added image to move them. 
\dsbase{} therefore measures v-distraction well and t-distraction
barely at all. This limitation is what motivated the two harder splits:
\dsname{}-Retrieved, whose text side is exam-grade reading comprehension
rather than captions that are easy by construction, and the hand-written
\dsname{}-Human.

\subsection{The Relation Between Accuracy and Distraction Rate}
\label{sec:asymmetry}
The text-over-vision asymmetry is \emph{not} universal. Because \dsbase{}'s text side is easy
by construction, we re-measure both directions across all seven models on \asmname{}, where T-only performance is no longer near ceiling. We define two metrics: i) a model's \textbf{grounding strength} in a modality is its accuracy from that modality alone, either V-only or T-only; and ii) its \textbf{distraction asymmetry} as its v-distraction minus t-distraction.

Results show that the less-grounded modality
is also the more distracted one (Table~\ref{tab:assembled}). The two
Qwen2.5-VL models show the classic text-over-vision pattern because their vision is the slightly
weaker side. The five others show the reverse, their text grounding being the weaker side.

A model's \textbf{grounding gap}, its T-only minus its V-only accuracy, correlates with its
distraction asymmetry at Pearson $r{=}{+}0.86$ over the seven backbones
In other words, models with higher base accuracy in a given modality are less likely to be distracted by the other modality. Dropping any single backbone and recomputing still yields $r$ always above $+0.79$, so the relation is robust to which models we evaluate. While this might seem obvious, note that a low accuracy in a modality does not necessarily imply a distraction. 
This is because distraction is measured only on the items a model already answers correctly from the grounded modality, so lower accuracy does not automatically mean more distraction.  In Section~\ref{sec:mechanism} we try to analyze potential causes of modality distraction.

\begin{table}[t]
    \centering\scriptsize
    \caption{Grounding, distraction, and their asymmetry on the audited \asmname{} (A-OKVQA
    Vision / RACE-High Text). Gap $=$ T Ground $-$ V Ground; Asymmetry $=$ V-Distr $-$ T-Distr. The CI is a per-model bootstrap over items. All values are percentages, and Gap and Asymmetry are
    differences in percentage points. Per-configuration counts and Wilson CIs: Table~\ref{tab:cell_cis}.}
    \label{tab:assembled}
\input{tables/assembled}
\end{table}

\begin{figure}[t]
\centering
\includegraphics[width=\linewidth]{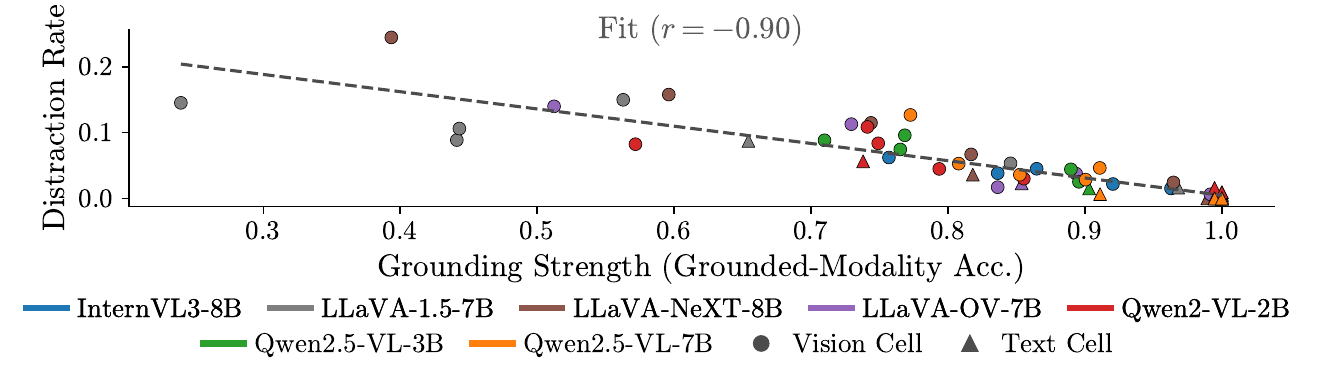}
\caption{The grounding-strength relation over all $70$
backbone$\times$domain$\times$modality configurations ($r{=}{-}0.90$, $95\%$ CI $[-0.97,-0.88]$ resampling model families). Each point is one backbone on one
domain in one modality, and the dashed line is one fit shared by every configuration.}
\label{fig:lawpair}
\end{figure}

\textbf{The grounding-strength relation.} We find that both patterns, namely that weaker backbones are
more distractible and that which modality gets distracted is model-dependent, follow one
empirical rule. Distraction scales inversely with grounding strength in the target modality. We empirically estimate distraction as follows:
\begin{equation}
\text{distraction} \;\approx\; a - b\cdot\text{grounding},\qquad b>0,
\label{eq:law}
\end{equation}
We estimate Equation~\ref{eq:law} with each datapoint as one backbone measured on one domain in one modality, giving $70$ configurations 
in total, $56$ on \dsbase{} and $14$ on \asmname{}. The constants $a$ and $b$ are \emph{shared
across all configurations}, fit once by least squares, leading to $a{\approx}27$, $b{\approx}0.26$. As a result, a single
line, not fitted per-model or per-domain, successfully predicts backbone, domain, and modality. Over all
$70$ configurations,
Pearson $r=-0.90$, and holds within vision configurations alone at $r=-0.82$ (Figure~\ref{fig:lawpair}, with additional robustness checks in Appendix~\ref{app:horserace}). 

A relation across models could still be confounded by what differs \emph{between} models (scale, training data, tokenizer). We exclude this with a
within-model ablation on \emph{every} backbone: holding the model fixed, we gradually lower its visual
grounding by downscaling the input image and re-measure v-distraction. The
resulting $84$ configurations ($7$ models $\times$ $2$ pools $\times$ $6$ levels) correlate at
$r{=}{-}0.91$ and lie on the relation line: artificially degrading grounding inside a fixed model reproduces the
across-model relationship. 

\textbf{Modality Distraction on Frontier Models.} The relation also extends beyond open $2$-$8$B models. We evaluate Gemini~3.5~Flash and Claude~Sonnet~5 under the same protocol on \dsname-Human,
where verification does not come from another oracle. Both land where the relation predicts:
they are the two best-grounded systems we measure (V-only $87\%$ and $95\%$ vs.\ $75\%$ open-model
average), among the least distracted (v-distraction $9.1\%$ and $11.7\%$ vs.\ $27.4\%$).
This shows that distraction is reduced for frontier models, not removed.

\subsection{Tracing Distractions}
\label{sec:mechanism}

From here on, we focus on vision-side distraction: irrelevant text makes the model change a correct answer it could give from the image alone. There are two reasons. First, it is the closer setting to deployed systems. Images routinely come together with text such as a caption or a retrieved passage. 
Second, it is the more frequent direction: on \dsname{}-Human it dominates on every backbone. We still measure the text-side cost of our interventions in Appendix~\ref{app:oneitem}.


\begin{figure}[t]
\centering
\includegraphics[width=\linewidth]{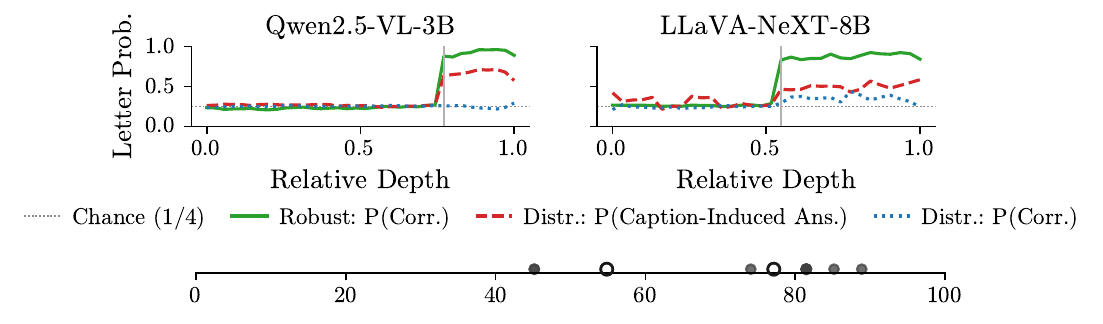}
\caption{Logit-lens trajectories: the option-letter probability recovered from each layer's
residual (softmax over the four answer letters, averaged over items) vs.\ relative depth.
Trajectories hover near the $1/4$ chance line until a mid-to-late ``commit'' layer (gray vertical), where robust vision items settle on the correct answer (green) and distracted items on the
\emph{caption-induced} wrong answer (red, dashed), their correct-answer probability (blue, dotted)
falling below chance. Bottom strip: commit depths of all backbones and pools; open markers are the
two backbones plotted above.}
\label{fig:lens}
\end{figure}

\textbf{Distraction is decided before the caption arrives.}
We find that which v-grounded items a caption will distract is decided by the model's answer
margin, even before the irrelevant caption is attached. For every item and every backbone, we run the same item
twice. Once with the image alone and once with both the image and the caption, holding the question,
the options, and the image fixed so that the presence of the caption is the only difference. From each run,
we read the logits of the four option letters. The answer \textit{margin} is the gap between the
logit of the correct option and the highest logit among the wrong options in the image-only run.
It is positive whenever the model answers the item from the image alone, and the smaller it is,
the less the logit the caption has to perturb to make the answer flip. Across backbones, the margin predicts distraction at AUROC
$0.87$ (\dsbase{}) and $0.92$ (\asmname{}), measured before the caption is seen. Thus, the confidence of a model, expressed as the gap between the top two answer options when given no irrelevant caption, already predicts how likey that its prediction will be distracted.

\textbf{Distraction is abrupt, not gradual.} The answer becomes confident abruptly at a
mid-to-late layer, and on distracted items the model commits there to the \emph{caption-induced}
answer rather than the correct one. The margin intuition tells \emph{which} items flip. We locate
\emph{where} in the forward pass the flip occurs with a logit
lens~\citep{nostalgebraist2020logitlens} over the per-layer residuals. Commitment falls at
$45$-$89\%$ depth across the backbones, with better-grounded models committing later (Pearson
$r{=}{+}0.88$ across backbones), as shown in Figure~\ref{fig:lens}.

\section{Mitigating Distraction}
\label{sec:mitigation}

In this section, we describe a method that trains a LoRA adapter for distraction reduction, which is trained on a split of our \dsbase{} data. 

 \textbf{Training setup for modality reduction.} We split \dsbase{} $60/20/20$ into train, validation and test ($2{,}050$, $683$ and $685$ items). \asmname{} is assembled by us and has no standard split, so we halve it once under a fixed seed, with the same fraction of domains and grounding labels in each portion. To measure what the robustness vector costs in general capability, we use four multimodal benchmarks, MMStar~\citep{chen2024mmstar}, MMBench~\citep{liu2024mmbench}, ScienceQA~\citep{lu2022scienceqa} and SEED-Bench~\citep{li2024seedbench}, taking the first $800$ questions of each. We then define three pools by role. i) The \emph{train} pool is the train split of \dsbase{}. ii) The \emph{parameter selection split}, used to validate the hyperparameter of our method, is \dsbase's validation split, the first half of \asmname{}, and questions $1$-$400$ of each benchmark. iii) The \emph{report side}, used once as a test set at the end, is \dsbase's test split, the other half of \asmname{}, questions $401$-$800$ of each benchmark, and all of \dsname-Human. 
 Appendix~\ref{app:splitsides} provides details on these data pools.

\begin{figure}[t]
\centering
\includegraphics[width=\linewidth]{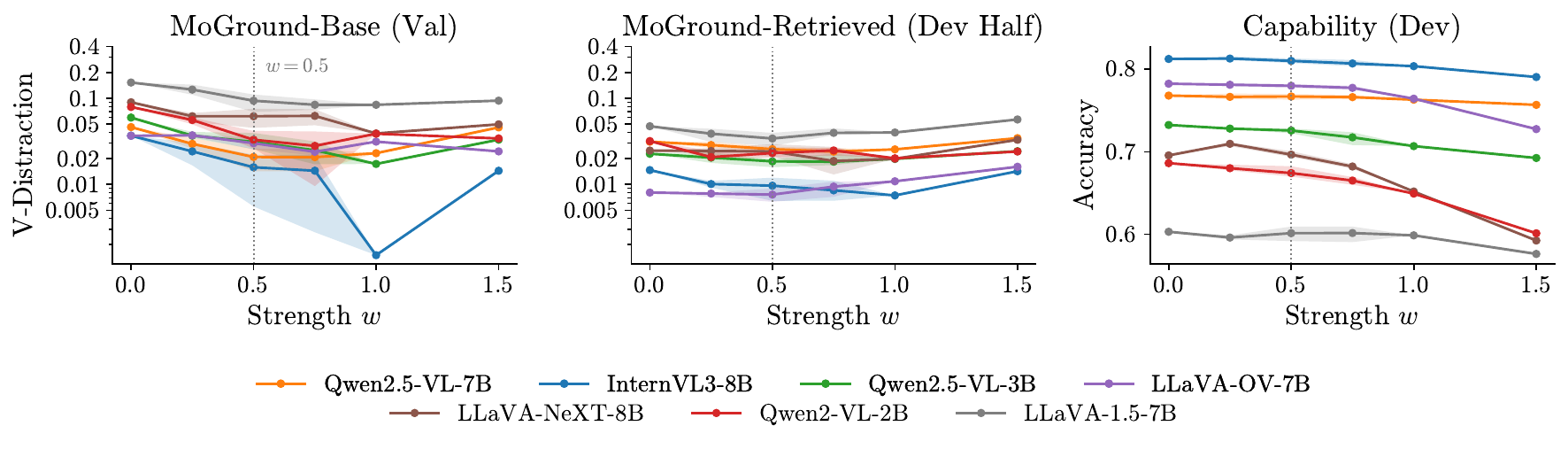}
\caption{Robustness vector performance on selection-side data as its strength $w$ varies. Bands span four training seeds, except for $w{\ge}1$.}
\label{fig:pareto}
\end{figure}

\textbf{LoRA-based tuning for distraction reduction.} The method is to fine-tune a LoRA adapter controlled by one number. The LoRA settings are fixed once and shared by all seven backbones:
rank $16$, $\alpha{=}32$, on the attention projections of the language model only. The only
quantity we tune is the strength $w$ at which the adapter merged with the base model. We fine-tune on \dsbase's train
split with a distraction-robustness objective. Vision-grounded items are answered with the caption
either dropped or replaced by an adversarial one, written by an oracle that sees the image and alters only the queried detail so that it points to the wrong option most easily confused with the right one.
Text-grounded items keep both inputs, which preserves text capability, and are anchored to the
base model with a KL term to contain side effects. The full pipeline is in
Appendix~\ref{app:recipe}. Multiplying the two LoRA matrices yields a single weight update which can be seen \emph{robustness steering vector} $\Delta$. Following task arithmetic~\citep{ilharco2022editing}, we multiply it by $w$ and add it to the base model's weights, so the strength is adjustable without retraining. We sweep $w \in \{0, 0.25, 0.5, 0.75, 1, 1.5\}$ on the selection side of the data
(Figure~\ref{fig:pareto}) and pick $w{=}0.5$. Two insights inform that choice. Distraction falls
to a minimum at $w{\in}[0.5,0.75]$ and regrows after it, while capability deteriorates
monotonically past that minimum, reaching $-10.3$pp in the worst model by $w{=}1.5$. We take the
$w$ that maximizes the mean v-distraction reduction on the \asmname{} dev half, the pool that
measures transfer rather than the trained domains: $w{=}0.5$, at a mean reduction of $19.1\%$ and
a mean dev capability cost within seed noise at $-0.36$pp.

\begin{table}[t]
\centering\scriptsize
\setlength{\tabcolsep}{3.5pt}
\caption{The robustness vector at the default $w{=}0.5$. Each value is the mean over four training seeds. Base is the v-distraction rate at $w{=}0$ and Reduction is relative to that base. Capability $\Delta$ is an absolute change in accuracy points. The average comes with a $95\%$ interval from
$4{,}000$ bootstrap draws that resample \emph{items} within each backbone}
\label{tab:pareto}
\input{tables/pareto}
\end{table}

\begin{table}[t]
\centering\scriptsize
\caption{Mean v-distraction and the delta relative to no intervention at all, averaged across four seeds. Capability is the change in general capability. Robustness vector is applied with $w{=}0.5$.}
\label{tab:method_comparison}
\input{tables/method_comparison}
\end{table}

Fixed once, the robustness vector is evaluated on the reporting
data pool (Table~\ref{tab:pareto}). V-distraction improves for every backbone on every dataset, by up to
$73\%$ on \dsbase{} test, $67\%$ on \dsname-Human and $39\%$ on \asmname{} held-out. Performance on the four benchmark we used to test degradation of existing skills
moves under $1$pp on six of the seven backbones, between $+0.9$ and $-0.5$pp. 
Distraction is a conditional rate, so the reduction converts into end-to-end accuracy
only where distraction is frequent. So, the gain is more significant on \dsbase{} and \dsname{}-Human where distraction is more frequent, and more negligible on \asmname{}. Results are reported in Appendix~\ref{app:accdeltas}). 


\textbf{Baselines for modality distraction.} Among all baseline methods we tried, the
robustness vector is the only one that reduces distraction on all three pools. We compare against
two approaches. We try \emph{prompting} in two different variants: a one-line instruction, prepended to the prompt,
which tells the model that the caption may be irrelevant and that it should judge from the image, and a zero-shot chain-of-thought~\citep{kojima2022large}, which asks the model to reason first and to end with a final answer (both prompts in Appendix~\ref{app:prompts}). 
We also test \emph{Contrastive decoding}~\citep{favero2024m3id}, which runs the model
twice on the same input, once with the image and once without, and pushes the answer toward the prediction when given the image. 
We try three additional variants of activation steering methods, one based on Sparse Autoencoders (SAEs) \citep{cunningham2023sae}, one on a logistic probe, and one on model residuals. They require training SAEs and additional interventions, so we could only apply them to a subset of the models. These additional comparisons are discussed in Appendices~\ref{app:abl_grid} and~\ref{app:oneitem}, with the SAE configuration in Appendix~\ref{app:sae_details}, and our LoRA fine-tuning method outperforms all alternative methods.



Table~\ref{tab:method_comparison} shows the main results.
\emph{Prompting does not work.} The instruction raises distraction by $+10\%$ and
$+4\%$ on \dsbase{} and \dsname{}-Human, while leaving \asmname{} unchanged ($-1\%$). Chain-of-thought
is worse on all three, $+57\%$, $+20\%$ and $+158\%$, and it is the only method here that also costs
capability, $-1.5$pp on average even when
its unparseable generations are excluded. Its largest single effect on distraction is on \asmname{}, where
LLaVA-OneVision goes from $5$ distraction events to $37$ once it reasons step by step. Overall, telling explicitly which modality to
prefer to is not an instruction a model follows, a result that follows findings from
\citet{zhang2025modalitypref}. 
\emph{Contrastive decoding comes closest.} It is the only prior method that reduces distraction at all, by $27\%$ on \dsbase{} test and $19\%$ on \dsname-Human, but nothing on \asmname{}. Overall, unlike our LoRA-based method, these methods either cost general capability or fail to consistently reduce distraction.



\section{Conclusion}
\label{sec:conclusion}

We introduce \dsname, a dataset for studying and mitigating modality distraction, whose questions are guaranteed to be answerable either with a caption or with an image, checked by a human audit. Modality distraction is a strict reliability loss because a model that has a correct answer becomes wrong when additionally given irrelevant context from another modality. Using \dsname{}, we found that modality distraction is not universal but depends on the grounding strength of a model, and it occurs abruptly in a mid-to-late layer. To mitigate the issue, \dsname{} enables a fine-tuning strategy that reduces distraction at negligible capability cost on average. 
We hope this work establishes per-item modality-grounding verification as a standard design principle, so that future multimodal models are evaluated not only on whether they can answer, but on which modalities they rely on when answering, and how easily those answers can be distracted. \dsname{} makes all three of these questions measurable.

\section*{Acknowledgements}
This work is supported by the MUR FIS2 grant n. FIS-2023-00942 "NEXUS" (cup B53C25001030001) and partly by Sapienza University of Rome via the Seed of ERC grant "MINT.AI" (cup B83C25001040001); and in part by the U.S. Army Research Office under Army-ECASE award W911NF-07-R-0003-03; the U.S. Department of Energy, Office of Science; the ARPA-H-SOL-24-101 program; the IARPA HAYSTAC Program; NSF Grants \#2205093, \#2146343, \#2134274, \#2441832; and CDC-RFA-FT-23-0069.

\section*{AI Use Statement}
Large language models are involved in this work in three roles.

i) As \emph{components of the data pipeline}, two oracles build \dsbase{}. \emph{Gemini~3~Flash} generates
and verifies the SemArt and ROCO items, \emph{Claude Sonnet~4.6} the DCI and VisText ones, each
source handled end-to-end by a single oracle so the cross-oracle check can
re-answer every item with the model that did not build it. Gemini~3~Flash also writes the
adversarial captions used in fine-tuning. The prompts are reported in
Appendix~\ref{app:oracle_prompts}.

ii) As \emph{subjects of study}, two proprietary models are scored alongside the seven open-weight backbones
in the frontier stress test, \emph{Gemini~3.5~Flash} and
\emph{Claude~Sonnet~5}. They are measured and never used to produce any other result.

iii) As a \emph{writing and coding assistant}, Claude Fable 5 and Opus 5 were used for code development, analysis
tooling, and to polish wording. The research questions, the experimental design, the analysis, and
the argument are from the authors. In this manuscript, most of the text is human-written, and every reported number is
produced by the released scripts and artifacts rather than by an LLM.

\section*{Ethics Statement}
\dsbase{} is built from four publicly available image-caption benchmarks (DCI, VisText, SemArt,
ROCO); the ROCO radiology images are de-identified, and we introduce no new personal data. The
only human annotation is by the authors (Appendix~\ref{app:validation}); no external annotators or
crowdworkers were involved. The work targets a reliability failure of VLMs, i.e., models answering correctly first but becoming wrong under irrelevant distraction. We foresee no misuse beyond the standard risks of any public benchmark: the biases of the source benchmarks and the possibility of training on them rather than using them for evaluation only.

\section*{Reproducibility Statement}
Everything behind the paper's numbers is released (Appendix~\ref{sec:release}): the verified
dataset with fixed splits and per-item three-condition records, the audited \asmname{},
the oracle prompts (Appendix~\ref{app:oracle_prompts}), the fine-tuning recipe
(Appendix~\ref{app:recipe}) and the selection/reporting protocol
(Appendix~\ref{app:splitsides}), and the scripts reproducing every
table and figure. One unavoidable reproducibility issue is that oracle models evolve, and the exact versions we use in this work might not be available in the future, so a practitioner is unable to regenerate the \dsbase{} exactly. 

\section*{Limitations}
\label{sec:limitations}
\paragraph{Scope of the claims.} Our measurements use multiple-choice answering by
open-weight $2$-$8$B vision-language models when \emph{answer-irrelevant} context is added to an
item the model already answers from one modality. The robustness vector survives a change of output format: with the
options removed and the same items answered in free-form generation, v-distraction still falls on
every backbone, by $39\%$ on the \dsbase{} test split, $24\%$ on \dsname-Human and $11\%$ on
\asmname{}. Each generated answer is mapped back to an option by embedding similarity, or by value
when the options are numbers. The distraction measurements, the grounding-strength relation and
the margin mechanism are nonetheless all measured in the multiple-choice setting. We do not test retrieval
or agentic pipelines, or adversarial or answer-bearing context, and
the grounding-strength relation is fitted over $70$ configurations from seven backbones, which is a small
sample for a scaling claim. Two proprietary frontier models are only scored on \dsname-Human
and extend the relation without breaking it. Beyond that, the leave-one-model-out forecasts test generalization
only across open backbones of this family and size range. Statements about what steering cannot do are similarly bounded by our setup, layers, and selection rules we tried.

Every \dsname{} item has its answer in exactly one modality, so we say nothing about
\emph{complementary} items whose answer needs both modalities together.
Absolute distraction is modest. On \dsbase{} it runs from $3.4\%$ to $13.6\%$ across backbones,
consistent with natural single-modality items rather than adversarially created ones. The
oracle filter might inherit the biases of the generating/judging models. We mitigate but do not eliminate
this via the three-condition filter, and the cross-oracle check
confirms that single-modality answerability is a property of the items rather than of the
verifying oracle, and two different oracles have a very high agreement rate. Finally, the generation-and-verification pipeline depends on proprietary oracles whose
versions change over time. The released dataset and its per-item records are fixed artifacts, but re-running the
pipeline would not reproduce them exactly.

\bibliographystyle{iclr2027_conference}
\bibliography{refs}

\appendix

\section{Released Artifacts}
\label{sec:release}
All artifacts are released under an open license to support reuse. 

The code is at \url{https://github.com/LuckerZOfficiaL/Modality-Distraction}, the data at \url{https://huggingface.co/datasets/LuckerZ/MoGround}, and the sparse autoencoders at \url{https://huggingface.co/LuckerZ/moground-saes}. Image files are not redistributed, since two of the six sources forbid it. The release lists every image with its upstream identifier and checksum, and ships a script that rebuilds the image folders and verifies them. The released artifacts are:

\begin{itemize}
 \item \textbf{\dsbase{}}, $3{,}418$ oracle-verified single-modality-grounding items
 (image, caption, MCQ, grounding label) with per-domain / per-grounding counts
 (Table~\ref{tab:composition}) and the fixed $60/20/20$ train/val/test split, together with the
 $250$ human-audit labels of Appendix~\ref{app:validation}, the seed that reproduces the audited
 sample, and the annotation notebook.
 \item \textbf{\dsname-Human}, the $125$-item hand-authored hard subset ($63$ vision- / $62$
 text-grounded across the four domains), written on seeds disjoint from \dsbase{} with captions
 designed to distract, and never used for any selection (\S\ref{sec:dataset}).
 \item \textbf{\asmname{}}, the $4{,}757$ audited items ($2{,}261$ A-OKVQA vision /
 $2{,}496$ RACE-high text) with cosine-retrieved cross-modal distractors and natural difficulty on
 both modalities from CC3M, for two-sided measurement of the distraction asymmetry.
 We release the post-audit set, the one every number in the paper is computed on, so that users
 need not re-derive which items to drop. The irrelevance judgments for the discarded items ship
 alongside it.
 \item \textbf{Oracle generation+validation pipeline}, code to replicate and run the three-condition-verification in new domains.
 \item \textbf{Vision-language SAEs} and their training code, for the two backbones they were
 trained on (Qwen2.5-VL-3B, LLaVA-NeXT-8B), with the configuration in Appendix~\ref{app:sae_details}.
 \item \textbf{Evaluation + analysis code}, the modality-robustness harness and the scripts
 reproducing every table/figure.
\end{itemize}

\section{Dataset construction pipeline}
\label{app:pipeline}
\dsbase{} is built in the following stages. All prompts and code are released.

\subsection{Three-condition verification}
\label{app:threecond}
The oracle then answers each candidate under three input
conditions, image-only ($V$), text-only ($T$), and image+text ($VT$), and we keep only items whose
answers match a single-modality signature:
\begin{itemize}
  \item \textbf{vision-grounded}: $VT$ correct $\wedge$ $V$ correct $\wedge$ $T$ wrong;
  \item \textbf{text-grounded}: $VT$ correct $\wedge$ $V$ wrong $\wedge$ $T$ correct.
\end{itemize}
Of $14{,}373$ generated candidates, $3{,}671$ are verified, the $74.5\%$ rejection rate quoted in
\S\ref{sec:dataset}. After deduplication and splitting this yields $3{,}418$ released items: DCI $1{,}613$, VisText $676$,
SemArt $497$, ROCO $632$ (Table~\ref{tab:composition}). Verification is \emph{operational}: single-modality answerability holds
relative to these oracles, not as absolute ground truth.

Each source is generated and verified by one oracle: Claude Sonnet for DCI and VisText, Gemini for
SemArt and ROCO. The cross-oracle check of \S\ref{sec:dataset} uses this split. Its $240$-item
sample takes $60$ items from each source, and every item is re-answered under all three conditions
by the oracle that did not verify it, Gemini re-answering the DCI and VisText items and Claude
Sonnet the SemArt and ROCO ones. The reported $94\%$ is the mean over the two labels of the rate at
which the second oracle answers the item correctly from its intended modality and from $V{+}T$
(vision $88.3\%$, text $99.2\%$).
\begin{table}[t]
    \centering
    \scriptsize
    \caption{Composition of our datasets: \dsbase{} (oracle-verified), \dsname-Human
    (hand-authored), and \asmname{} (established VQA and reading-comprehension benchmarks). For \dsbase{}
    the Candidates and Kept columns give the candidates generated per source and the survivors of
    the three-condition filter. Kept counts precede deduplication and id renumbering, so
    they sit slightly above the released V-grd/T-grd totals. \asmname{} counts are post-audit
    (Appendix~\ref{app:audit_retr}).}
    \label{tab:composition}
    \setlength{\tabcolsep}{3.0pt}
\input{tables/composition}
\end{table}

\subsection{Oracle prompts}
\label{app:oracle_prompts}
The templates below are the ones that produced \dsbase{}, reproduced from the released code so
that the generation and verification passes can be re-run or audited. The final pair produced the
adversarial captions used only as fine-tuning signal in \S\ref{sec:mitigation}, never for any
released item.

\input{tables/appendix_oracle_prompts}

\paragraph{Adversarial captions (mitigation fine-tuning), system prompt.}
\begin{tcolorbox}[breakable,colback=gray!5,colframe=gray!45,arc=1mm,boxrule=0.3pt,width=\textwidth,
                  top=3pt,bottom=3pt,left=4pt,right=4pt,fontupper=\ttfamily\scriptsize]
\raggedright
You write image captions for a research dataset. Given an image, a multiple-choice question\\
about it, the options, and which option is CORRECT, you fabricate a fluent, natural-sounding\\
caption of the SAME scene that would mislead a careful reader into choosing a specified WRONG\\
option. The caption must stay consistent with everything else in the scene and only alter the\\
single detail the question is about. Never hedge, never say the caption is wrong, and never use\\
self-referential words like 'image', 'photo', 'caption', 'shown', or 'according to'.
\end{tcolorbox}

\paragraph{Adversarial captions, per-item message.} Sent with the item's image attached.
\begin{tcolorbox}[breakable,colback=gray!5,colframe=gray!45,arc=1mm,boxrule=0.3pt,width=\textwidth,
                  top=3pt,bottom=3pt,left=4pt,right=4pt,fontupper=\ttfamily\scriptsize]
\raggedright
QUESTION: \{q\}\\
OPTIONS:\\
\{opts\}\\
CORRECT OPTION: \{gold\_letter\}. \{gold\_text\}\\
\ \\
TRUE CAPTION (accurate; for scene grounding only):\\
\{cap\}\\
\ \\
Task: pick the WRONG option that is most plausibly confusable with the correct one, then write\\
ONE adversarial caption (roughly \{nchar\} characters, similar in style and length to the true\\
caption) that describes this scene as if that wrong option were true. Keep every other detail\\
faithful; only the detail the question asks about should point to the wrong option.\\
Return JSON: \{"target\_index": <int 0-based of the wrong option you argued for>,\\
"adv\_caption": "<the caption>"\}
\end{tcolorbox}

\paragraph{Splits, and what the analyses use.} We release a fixed $60/20/20$ train/val/test split
(train $2{,}050$ / val $683$ / test $685$, with the same fraction of domains and grounding labels in each part) for \emph{downstream
supervised} uses such as training probes or steering vectors. The \emph{diagnostic} analyses
(\S\ref{sec:leaderboard}-\S\ref{sec:mechanism}) do \emph{not} partition by this split: they only
read frozen-model behavior and activations, so the benchmark, the grounding-strength relation, and the
mechanistic analyses are computed over the \emph{entire} dataset, and the predictive claims are
held out by construction (leave-one-configuration-out and leave-one-model-out, \S\ref{sec:asymmetry}). The
one place \dsbase{} is trained on is the mitigation (\S\ref{sec:mitigation}): the robustness
vector is fine-tuned on the train split only, its \dsbase{} effect is reported on the held-out test
split, and its transfer on \asmname{}, which no fine-tuning ever touches. Every frozen-model
result, for all seven backbones, is computed on this full oracle-verified set. We never restrict
one backbone's evaluation to items selected by another model.

\section{Example \dsbase{} and \asmname{} items}
\label{app:examples}
For each \dsbase{} domain we show one vision-grounded and one text-grounded item. For \asmname{}
pool we show one A-OKVQA (vision) and one RACE-high (text) item. In every case the answer is
grounded in exactly one modality while the other modality is answer-irrelevant, and the correct option
is set in \textbf{bold}. We additionally show four \dsname-Human items, which are hand-authored under the same design rules: their captions are written to \emph{prime} a wrong
option without answering the question. The mechanism is visible in each. The painting's caption
mentions the sun, an eagle and pigeons, three of the four options, each attached to a different
referent (the city's usual weather, the symbolism of the birds), leaving the correct answer the
only option it never names. The chart's caption supplies four other figures including a value for
the queried year in a different region. The hat photo's caption enumerates the colors present as
``white, blue, green, beige, black'', naming three wrong options and omitting the right one. Each
of these items distracts several backbones.
\input{tables/appendix_examples}

\section{Additional dataset validation}
\label{app:validation}

\subsection{Human audit of \dsbase{}: single-modality answerability}
\label{app:audit_base}
The three-condition gate and the
cross-oracle replication of \S\ref{sec:dataset} both verify \dsbase{} by machine. To check that
the verified property aligns with human perception, two of the authors independently
audited the same random sample of \dsbase{}.

\emph{Sample.} $250$ items, $125$ vision-grounded and $125$ text-grounded, drawn across all four
domains in proportion to their sizes. The draw is a fixed permutation of the corpus under seed
$20260804$, so both the sample and the order in which it was presented are reproducible from the
released code. Each annotator judged the items in that order, with no opportunity to select which
ones to judge.

\emph{Protocol.} For each item each annotator saw the image, the caption, the question and the
four options, in the same form a model receives them, and recorded a pass/fail verdict together
with a reason on failure. An item passes only if both halves of the guarantee hold: the intended
modality answers it, \emph{and} the other modality does not. No model predictions were shown at any
point, so the judgments are independent of how any backbone behaves on the item. The only additional tool available for the annotators is the ability to zoom in on the image.

\emph{Result.} Both annotators confirm the guarantee on almost every item. An item passes only if
the annotator raised no objection of any kind, including defects that have nothing to do with modality grounding. On that strict reading $234$ and $232$ of the $250$ items pass ($93.6\%$
and $92.8\%$), and the two annotators agree on $236$ of them (Table~\ref{tab:audit}). Among rejections, most have nothing to do with modality grounding. Of the first annotator's $16$
rejected items, ten are items where the option marked as correct is not the right answer in the
first place. The other six are grounding failures: four where the intended modality does not
answer the question after all, and two where the other modality answers it as well. Counting only
those grounding failures, the guarantee holds on $97.6\%$ and $96.8\%$ of the sample, and the
leakage the filter exists to prevent appears on just $2$ and $4$ items. Per-domain pass rates run from $90.9\%$ to $96.7\%$ for the first annotator and $86.4\%$ to $98.4\%$ for the second, so no single domain accounts for the rejections. Both label sets ship with the release, together with the sampling seed and the annotation notebooks, so the audit can be re-examined.

\begin{table}[t]
\centering\small
\caption{The human audit, performed independently by two annotators on the same items, in the same
order, under the same protocol. The two agree on 236/250 \dsbase{} items and 189/200
kept \asmname{} items. The annotators' instruments split the non-leak categories differently
(annotator 1 separated wrong-option support from out-of-set answers. Annotator 2 folded both into
conflict), so cross-annotator comparison is at the leak / conflict / defect level. Leak counts are
directly comparable.}
\label{tab:audit}
\input{tables/audit_tworater}
\end{table}

\emph{Distraction re-evaluated on verified items only.} The audit's $6.4\%$ failure rate is of the same order as the smallest distraction rates we report, so we re-compute v-distraction on the audited vision items, keeping only the
$112$ items a human verified. Distraction survives on every backbone. Three rates rise and four fall, by
at most $1.7$pp, which is how sampling noise on $49$-$97$ items behaves rather than the
one-directional drop that verification errors would produce.


\subsection{Human audit of \asmname{}: irrelevance of the added context}
\label{app:audit_retr}
Retrieval by embedding similarity makes a
distractor topically plausible but cannot guarantee it is answer-irrelevant. We therefore audit
every retrieved caption in \asmname{} with an LLM judge, in two independent passes with
$95\%$ agreement on a $120$-item overlap. $9.6\%$ of captions violate irrelevance: $5.8\%$ assert
a wrong option and $3.7\%$ state the correct one. All violating items are dropped, and every
behavioral \asmname{} statistic in the paper is computed on the remaining clean subset ($4{,}757$
items).

To check the judge itself, the same two annotators hand-audited $250$ pool items under the
protocol above: a fixed permutation, here under seed $20260818$, judged in the order presented,
with no model predictions shown. Only the question being asked is different. Here it is about the \emph{added} irrelevant-modality content, and each annotator marks whether that content answers, supports, or contradicts one of the four options. The sample covers three
pre-specified subsets: $140$ kept vision items, $60$ kept text items, and $50$ of the $239$ items
the judge dropped. Some kept items have ambiguous options ($10$ and $7$ found by the two annotators). That is a fault in the source benchmarks, not in our pipeline. We exclude those items from the results below.

Of the usable kept items, $187/190$ and $189/193$ are answer-irrelevant ($98.4\%$ and $97.9\%$),
the two annotators agreeing on $94.5\%$ of kept items (Table~\ref{tab:audit}), and the released pool is essentially
leak-free: one genuine leak in $383$ usable judgments between them ($1$ and $0$), and none of the
$60$ images added to text items leaks for either annotator. On the items the judge dropped, the
annotators confirm $50\%$ and $67\%$ of usable rejections as leaking, conflicting, or pointing
outside the option set ($18$ and $10$ genuine leaks). Leaks thus concentrate almost entirely in the
pool the judge dropped, so the filter genuinely separates modality leaks from non-leaks rather than dropping
arbitrarily.

\emph{What the remaining failures are.} None of the other flagged items involves leakage. On
\dsbase{} they are faults inherited from the source datasets: items whose marked answer is wrong
($10$ and $9$) and one image-caption conflict. On the \asmname{} side they are captions that \emph{contradict} an
option rather than answer it ($2$ and $4$ kept items), behaving as cross-modal conflict rather
than as irrelevant context. This failure appears only in \asmname{}, because \dsbase's captions are generated under a no-conflict constraint.

\subsection{Agreement between the annotators}
\label{app:audit_agree_sec}
Collapsing each annotator's labels to leak vs.\ no
leak, the two agree on $246/250$ \dsbase{} items ($98.4\%$) and on $199/200$ kept \asmname{} items
($99.5\%$). Table~\ref{tab:audit_agree} gives the full breakdown.
\begin{table}[t]
\centering\small
\caption{Leak / no-leak agreement between the two annotators: \dsbase{} sample (left,
$n{=}250$) and the kept half of \asmname{} (right, $n{=}200$; dropped items excluded).
Agreement is $98.4\%$ and $99.5\%$.}
\label{tab:audit_agree}
\input{tables/audit_agree}\hspace{2em}%
\begin{tabular}{lcc}
\toprule
\asmname{} (kept) & A2: leak & A2: no leak \\
\midrule
A1: leak & 0 & 1 \\
A1: no leak & 0 & 199 \\
\bottomrule
\end{tabular}
\end{table}

\subsection{The open-weight backbones}
\label{app:backbones}
The seven models are built from different parts, but they all consist of a language model and an
image encoder. The language models are Vicuna-7B for LLaVA-1.5-7B, Llama-3-8B for LLaVA-NeXT-8B,
Qwen2-7B for LLaVA-OneVision-7B, Qwen2.5-7B for InternVL3-8B, and Qwen models for the three
Qwen-VL models. Regarding the image encoder, two older LLaVA models use CLIP and receive the image
at $336$ pixels. LLaVA-OneVision uses SigLIP at $384$ pixels. InternVL3 uses its own image encoder
at $448$ pixels. The Qwen-VL models have no fixed input size.

They also receive images in different ways. LLaVA-1.5 shrinks every picture into one small square.
LLaVA-NeXT and LLaVA-OneVision cut a large image into several smaller ones and process each of
them. InternVL3 does the same but then shrinks each small patch further. The Qwen-VL models
receive the image at whatever size, so a bigger image simply becomes more tokens.

What they all do the same way is the cross-modal fusion. Each one turns the image into tokens that
the language model treats like words, and places the image tokens in the same sequence as the text
tokens.


\subsection{Per-domain breakdown}
\label{app:perdomain}
\begin{figure}[t]
\centering
\includegraphics[width=\linewidth]{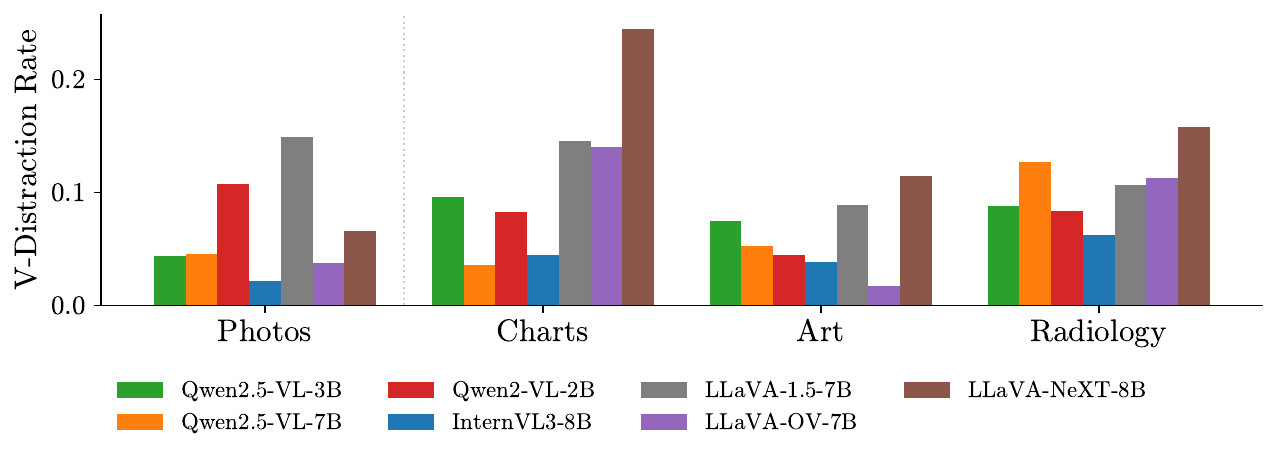}
\caption{Per-domain v-distraction for all seven backbones, behind
Table~\ref{tab:leaderboard}: an answer-irrelevant caption distracts vision answering more on
non-natural domains (charts, art, radiology) than on photos in five of the seven backbones
(exceptions discussed in the main text).}
\label{fig:distraction}
\end{figure}

Figure~\ref{fig:distraction} plots the per-domain v-distraction
columns of Table~\ref{tab:leaderboard} for all seven backbones. Table~\ref{tab:tdistr_base} does the
same for the text side, where the rate is at or near zero in every domain.
\begin{table}[t]
    \centering\small
    \caption{Text-grounded half of Table~\ref{tab:leaderboard}.
    All values are percentages. T$+$V and T-only are accuracy on text-grounded items with both
    inputs and with the caption
    alone. T-Distraction is the fraction of the T-only-correct items that the added image flips,
    per domain and pooled. The rate is at or near zero in every configuration, and they are well
    powered ($n{=}182$--$557$ solved items each), so the ceiling is a property of the
    verified text side, not of any one domain.}
    \label{tab:tdistr_base}
\input{tables/tdistr_base}
\end{table}


\subsection{Counts and confidence intervals}
\label{app:cellcounts}

 Table~\ref{tab:cell_cis} gives the counts behind every distraction rate, with a $95\%$ interval for each.

\begin{table}[h]
\centering\small
\caption{Counts and Wilson $95\%$ intervals behind each distraction rate. Each entry reads \emph{flipped/solvable}: how many items the model answers correctly from the grounded modality alone (solvable), and how many of those it then answers wrongly once the other modality is added (flipped). Top block: \asmname{}. Bottom block: \dsbase{} with domains pooled.}

\label{tab:cell_cis}
\input{tables/appendix_cell_cis}
\end{table}

\subsection{Both directions of the caption's effect}
\label{app:flips_sec}
Table~\ref{tab:flips} splits the caption's effect on v-grounded items into its two directions: it breaks items the
image alone answers, and it fixes items the image alone misses. Both happen on every backbone, but
breaking is the dominant event, so V$+$T accuracy sits below V-only accuracy everywhere.

\begin{table}[t]
    \centering\small
     \caption{Both directions of caption-induced answer change on the $n{=}2067$ vision-grounded items of \dsbase{}. Broken counts items the model answers from the image alone but misses once the caption is added, and Fixed counts the reverse. V-Distraction is Broken over the image-answerable items, Recovery is Fixed over the rest. All values are percentages except the two counts.}
    \label{tab:flips}
\input{tables/flips}
\end{table}

\section{Additional analyses}
\label{app:analyses}

\subsection{Does anything else predict distraction?}
\label{app:horserace}
Grounding strength predicts distraction, and it predicts a held-out backbone's rates to within
$1$-$2$pp on average, but a predictor is only interesting if simpler ones do worse. We compare it against
eight alternatives on the same $70$ configurations. All of them use the same forward
pass, so none costs extra data. We score each by its correlation with distraction and by its error
when predicting a backbone left out of the fit, which is the only number that says how good a predictor on an unseen model.

\begin{center}\scriptsize
\begin{tabular}{llccc}
\toprule
Predictor & What it measures & Access& $r$ & Error on a held-out model (pp) \\
\midrule
\textbf{Grounding strength} & Single-modality accuracy & Black box& $-0.899$ & $\mathbf{1.60}$ \\
\midrule
Answer Confidence& Mean max softmax& Logits& $-0.901$ & $1.65$ \\
Answer Entropy& Mean softmax entropy& Logits& $+0.887$ & $1.84$ \\
\midrule
Question Length& Mean words per question& Black box& $+0.589$ & $3.15$ \\
Calibration Error& Calibration gap over $10$ bins & Logits& $+0.665$ & $3.70$ \\
Distractor Strength& Accuracy from the other modality alone& Black box& $-0.220$ & $3.85$ \\
Context Length& Mean words per caption or passage& Black box& $-0.147$ & $3.96$ \\
Model Scale& $\log_{10}$ parameter count & Metadata& $+0.085$ & $4.05$ \\
\bottomrule
\end{tabular}
\end{center}

Two of the eight alternatives matter. \textbf{Model scale does not predict distraction} ($r{=}{+}0.085$), so
distractibility is not something larger models escape. \textbf{Distractor-channel strength does not
either} ($r{=}{-}0.220$): what matters is the weakness of the grounded side, not the strength of
the competing one. Answer confidence and entropy do predict about as well, since they read the
same logit gaps, but grounding strength needs only whether the answer was correct, so it works on
models that expose no logits. The relation is also not an artifact of the obvious confounds. Question length, context length, option count, and how well the other modality answers all vary alongside grounding strength, but removing their influence moves the correlation only from $-0.90$ to $-0.84$. Taken one slice at a time, by source dataset, by modality, and on the ceiling-free pool, it still stays between $-0.72$ and $-0.99$.

\subsection{Distraction is not an artifact}
\label{app:placebo}

A conditional flip rate could in principle reflect generic instability under a longer prompt
rather than anything about the caption itself. We test this with three control captions per
vision item, graded by how much of the true caption's structure they preserve: a \emph{mismatched}
caption (another item's caption from the same domain, length-matched by construction: median
word-count difference zero), a \emph{scrambled} caption (the item's own words, order shuffled),
and \emph{neutral} filler (a fixed content-free sentence tiled to the caption's
length\footnote{The neutral caption: ``\textit{Conditions at the time were considered fairly
ordinary overall, and observers generally agreed that the situation developed in a way most would
describe as typical for the period in question, with nothing unusual reported.}'' It is repeated
end-to-end and truncated to each item's caption word count.}). Each
replaces the true caption in the V+T condition. All rates are computed on the identical denominator,
the items that backbone solves from V-only, so every contrast is paired.

\begin{table}[t]
\centering\small
\caption{Control captions. Conditional flip rate on each backbone's V-only-correct
items ($n$) when the added text is the item's true caption (True) vs.\ a length-matched
caption of another same-domain item (Mismatched), the item's own caption with word order shuffled
(Scrambled), or content-free filler of the same length (Neutral). Every contrast is paired on the
same items. The true caption is the strongest distractor for all seven backbones.}
\label{tab:placebo}
\input{tables/placebo}
\end{table}

Table~\ref{tab:placebo} reports the flip rate under the true caption and under each
control caption. The true caption is the strongest distractor on every backbone: it out-flips the scrambled
control on $7/7$ backbones and the content-free control on $7/7$, and the ranking [true caption $>$ \{mismatched, scrambled\} $>$ neutral caption] holds
strictly on all seven backbones. The ordering reflects semantic content. The true captions in \dsname, being fluent and relevant descriptions of \emph{this} scene, have more distractive power.

\subsection{SAE-feature ablation and patching}
\label{app:abl_grid}

Table~\ref{tab:ablation} gives the full grid behind the claim in \S\ref{sec:mitigation}. Every edit is applied only to distracted items, the ones the model answers correctly from the image alone and wrongly once the caption is added, and only at the last token of the listed layer.

We try three edits, each against its own random control. \emph{Removing SAE features} subtracts the $K$ features that fire most on distracted items relative to robust ones; the control removes $K$ features drawn at random from those that ever fire. \emph{Removing one direction} fits a probe to separate distracted from robust items and removes the single direction it uses; the control removes a random direction. \emph{Patching in the image-only state}, the positive control, replaces the residual with the same item's stored image-only residual; the control uses a different item's residual.

Removing SAE features recovers no more than removing random ones, at every $K$ and on both backbones, and removing the probe direction does no better. Patching the item's own image-only residual recovers $93.5\%$, against $24.3\%$ when a different item's residual is used. The site is therefore sufficient. What the edit needs is information specific to the item, not a direction shared across items.

\begin{table}[t]
\centering
\small
\caption{Controllability grid on distracted vision items, both SAE backbones. Recovery is the share of distracted items an edit fixes. The Null column applies the same edit at random. Robust Kept is the share of already-correct items left unchanged.}
\label{tab:ablation}
\input{tables/ablation}
\end{table}

\subsection{All interventions on one item set}
\label{app:oneitem}
Tables~\ref{tab:baseline}-\ref{tab:baseline_cap} score every intervention the same way, so
prompting, contrastive decoding, the activation edits and the robustness vector can be read side by
side. Each backbone is judged only on the items it answers correctly from the grounded modality
alone, the rule used everywhere else in the paper. Table~\ref{tab:ablation} instead scores each
activation edit the way it was designed, on distracted items only, so its numbers are not
comparable with the rest.

\begin{table}[t]
\centering\small
\caption{Every intervention on the two backbones that have SAE, on the \dsbase{} test split. ``Deployable'': whether a system could run the intervention knowing only the image, the context text, and the question. $\Delta$ V-Distr is the relative change in v-distraction from base (negative $=$ reduced). $\Delta$ Text is the change in text accuracy in percentage points. Both backbones are scored on identical items on the held-out reporting side.}
\label{tab:baseline}
\input{tables/baseline_dm}
\end{table}
\begin{table}[t]
\centering\small
\caption{The same interventions on the \emph{assembled} held-out pool, the cross-domain surface. ``Deployable'': whether a system could run the intervention knowing only the image, the context text, and the question. $\Delta$ V-Distr is the relative change in v-distraction from base (negative $=$ reduced). $\Delta$ Text is the change in text accuracy in percentage points. Both backbones are scored on identical items on the held-out reporting side (Appendix~\ref{app:splitsides}). \dsbase{} test is Table~\ref{tab:baseline}.}
\label{tab:baseline_asm}
\input{tables/baseline_asm}
\end{table}
\begin{table}[t]
\centering\small
\caption{General-capability cost of the three interventions that alter the served model, as the four-benchmark mean accuracy change in percentage points (report-side data). The activation steering methods
of Tables~\ref{tab:baseline} and~\ref{tab:baseline_asm} are applied only to items already known
to be distracted, and are not models a system could serve, so they have no capability cost to
report.}
\label{tab:baseline_cap}
\input{tables/baseline_cap}
\end{table}
Read the signs rather than the decimals. Each per-backbone configuration rests on $20$-$27$
distracted items, and a selection-side replication of the table agrees in every sign.
Chain-of-thought is scored on the answers the parser could read, and it produces no parseable
answer on $2.1\%$ (Qwen2.5-VL-3B) and $8.6\%$ (LLaVA-NeXT) of capability questions. The
interventions are listed below, with \checkmark for deployable and $\times$ for not.

\begin{itemize}
\item \textbf{Prompt instruction} (\checkmark). One sentence prepended to the user message telling
the model to judge from the image and use the caption only if it helps (prompt reported in
Appendix~\ref{app:prompt_instr}). Deployable: it is a fixed string, requires nothing about the item, and
adds a handful of tokens.

\item \textbf{Zero-shot chain-of-thought} (\checkmark). A fixed instruction to reason before
answering (Appendix~\ref{app:prompt_cot}). Deployable, but not free: it adds up to $256$ decoded
tokens on every query.

\item \textbf{M3ID contrastive decoding} (\checkmark). The published decoding rule of
\citet{favero2024m3id}. The model is run twice on the same item, once with the image and the text,
giving logits $l_{VT}$, and once with the text alone, giving $l_T$. The gap $l_{VT}-l_T$ is the
part of the prediction the image is responsible for, and the rule pushes the answer further along
it,
\begin{equation*}
l^{*} = l_{VT} + \mu\,(l_{VT} - l_T) \quad\text{if}\quad \max_k \mathrm{softmax}(l_{VT})_k < \alpha,
\qquad\text{and}\qquad l^{*} = l_{VT} \ \text{ otherwise.}
\end{equation*}
The condition is a confidence gate. When the model is already confident, its answer is kept, and
only unsure predictions are pushed toward the image. The strength $\mu \ge 0$ sets how hard to
push, and $\mu{=}0$ is ordinary decoding. Deployable, but not free, since the image-free run costs
a second forward pass on every query.

We change two things. M3ID was built for open-ended captioning, where $\mu$ grows with token
position as the image's influence fades over a long description. A single answer letter gives that
schedule nothing to act on, so we sweep $\mu$ as a constant instead. And the confidence gate is
computed over the four answer letters rather than the full vocabulary. We choose $(\mu,\alpha)$ on
the selection side, taking the pair with the largest v-distraction reduction among those that cost
no more than $0.5$pp of overall accuracy on either selection pool. Of the related decoding methods,
only M3ID maps onto our setting. VCD~\citep{leng2024vcd} contrasts against a noised image, which
targets hallucination from the visual side rather than from added text, and
OPERA~\citep{huang2024opera} penalises over-trust inside beam search over a generated sequence,
which has no analogue in a one-token forced choice.

\item \textbf{SAE feature ablation} and its \textbf{random-feature control} ($\times$). The top-$K$
distraction features are removed from the residual. Not deployable, for two independent reasons.
The protocol applies the edit only to items already known to be distracted, a label no system has
at inference time. And, more decisively, the edit does no better than removing the same number of randomly chosen
features, even when handed that label, so there is nothing to deploy.

\item \textbf{Single dense direction} ($\times$). The layer's most distraction-predictive direction
is projected out. Not deployable for the same two reasons as the SAE ablation. It is applied only to
items already known to be distracted, and it fails against its random baseline regardless.

\item \textbf{$h_V$ patch} ($\times$). The item's own image-only residual is substituted at the
commit layer, on distracted items only. This one \emph{works}, removing $75$-$100\%$ of
distraction across both pools, which is what makes it the paper's positive control. It is not deployable because the
gate is an oracle: it must be told which items are distracted, and that is precisely what a
deployed system does not know. It also needs a second, image-only forward pass to obtain $h_V$.

\item \textbf{$h_V$ patch, applied to every item} ($\times$). The same edit applied to every item, which removes
the oracle requirement. It keeps the distraction reduction and is therefore label-free, but it
overwrites the caption on text-grounded items too, where the answer lives: text accuracy falls by
$27$-$70$pp. Not deployable because the text accuracy drop is unacceptable.

 \item \textbf{$h_V$ patch, fired by a learned detector} and its \textbf{random-firing control} (\checkmark). Instead of being told which items are distracted, we train a detector to predict. A logistic probe reads the model's own state at the intervention layer, is fitted on \dsbase's train split, and the patch is applied to the items it scores highest, on as many items as the oracle would have chosen. This version is deployable, since the probe needs nothing but the model, and it is the one a practitioner would build. It repairs little. The probe tells distracted from robust items at AUROC $0.65$-$0.85$, but the patch then removes only $2$-$22\%$ of the distraction, against $75$-$100\%$ when told which items to fix, and barely more than patching the same number of items at random.

\item \textbf{Robustness vector} (\checkmark). A scaled weight update applied once, offline. Deployable in the strongest sense: after merging there is no gate, no
extra forward pass, and no added tokens, so inference is byte-identical to the base model.
\end{itemize}

\paragraph{The text side measured the same way.} Every method that reduces vision-side distraction
raises text-side distraction a little, and the robustness vector's increase is the smallest among
the methods that reduce it. Tables~\ref{tab:baseline} and~\ref{tab:baseline_asm} report the text
side as accuracy, which can move differently from a conditional rate. On LLaVA-NeXT-8B the
robustness vector costs $1.0$pp of \asmname{} text accuracy while its t-distraction rises by
$2.4$pp from a base of $3.0\%$. Table~\ref{tab:tdistraction} therefore measures both sides the same
way. On Qwen2.5-VL-3B the vector cuts v-distraction by $20\%$ for $+0.3$pp of t-distraction, where
prompting cuts $14\%$ for $+1.0$pp. On LLaVA-NeXT-8B it is the only method that reduces
v-distraction at all. We measure this on \asmname{}, because the text side of \dsbase{} test sits
at ceiling.

\begin{table}[t]
\centering\small
\caption{Benefit and cost of the four deployable interventions on the \asmname{} held-out pool, both measured as conditional rates. Vision side: relative change in v-distraction, negative is better. Text side: change in t-distraction in percentage points, positive is worse. Base Rate gives each backbone's v- and t-distraction before any intervention.}
\label{tab:tdistraction}
\input{tables/tdistraction}
\end{table}

\paragraph{Why there is no modality-classifier gate in the table.} An apparent way to make
the $h_V$ patch work is to apply patching on items a classifier predicts are vision-grounded. We do not
evaluate this as a baseline, because it cannot be validly trained on \dsbase{}: the modality
label is readable from surface text \emph{by construction}, text-grounded captions carry an
injected fact, and question categories differ by modality, and a bag-of-words classifier on the
question text alone already separates V from T at AUROC $0.989$ (accuracy $95.7\%$). Any modality
classifier fitted to \dsbase{} therefore learns the dataset's surface cues rather than image use, and a score on our pools would measure cue detection, not deployability. \asmname{}
pool's V/T label is likewise readable from the text, as dataset provenance. \dsbase{} verifies items, it
does not supply a training distribution for modality classification.

\paragraph{Is the robustness vector overfit to \dsbase{} too?} It is trained on \dsbase{}, so the
question is fair, and three measurements say it is not. First, the vector makes no per-item
prediction at inference, so there is no classifier that could rely on cues specific to
\dsbase{}. Second, it reduces v-distraction by $29\%$ on the \asmname{} held-out pool, which it
never saw in training, at a mean capability cost of $0.1$pp on four benchmarks that carry no \dsbase{} text. Third, its strength $w$ was chosen on \asmname{} dev rather than on any \dsbase{} pool. Overfitting would show as \dsbase{} validation improving while
\asmname{} gets worse, and no backbone shows that at $w{=}0.5$).

\subsection{Fine-tuning recipe (the robustness vector)}
\label{app:recipe}
Each backbone is fine-tuned with a LoRA adapter on attention only, rank $16$ and $\alpha{=}32$, on the $q/k/v/o$ projections of the language model. The vision tower is untouched. Training runs for $2$ epochs over \dsbase's $2{,}050$-item train split, with learning rate $10^{-4}$, batch size $1$, gradient accumulation $8$, and cross-entropy on the answer token only.

What the model sees depends on the item's grounding label. A vision-grounded item has its caption dropped half the time to encourage vision usage. The other half, the caption is made adversarial as often as it is truthful. A text-grounded item always keeps both the image and the caption, exactly as at test time, double-weighted in the loss to offset there being fewer such items, and a KL anchor (weight $2.0$) on the answer-token distribution against the frozen base model to prevent it from drifting too much. The dual strategy keeps the model using captions when the answer is in them, while the vision items teach it to doubt them.

The adversarial captions are written by Gemini 3 Flash, with the image in context. Given the question, the options, the correct answer, and the true caption, it picks the wrong option most easily confused with the correct one, then writes a caption of the same scene that changes only the
queried detail so that it points to that option. Length and style match the true caption (prompt in
Appendix~\ref{app:oracle_prompts}). This covers $1{,}185$ of the $1{,}252$ vision-grounded train items ($94.6\%$). The rest never take the adversarial branch, and the targeted option is never the correct answer.

For each backbone, we train four LoRA adapters across four seeds. The seeds differ only in which items have their caption dropped, which of the rest get an adversarial caption rather than a truthful one, and the order of the training examples.

\subsection{Selection and reporting splits}
\label{app:splitsides}
Every pool is split once, and the two halves do different jobs. The \emph{selection side} is where
we make choices, such as the strength $w$. It is \dsbase's validation split ($683$ items), half of
the audited \asmname{} pool ($2{,}378$ items), and the first $400$ questions of each capability
benchmark. The \emph{reporting side} is used only for final evaluation. It consists of
\dsbase's test split ($685$), the other half of \asmname{} ($2{,}379$), and capability questions $401$-$800$, which no selection has seen.

The capability questions are taken in the order the benchmark publishes them, with no reshuffling,
so both halves are fixed and reproducible. The exception is SEED-Bench, where we first drew
$3{,}000$ image questions with a fixed seed and then kept them in source order. The two halves of
\asmname{} are drawn to hold the same fraction of vision- and text-grounded items.

Note that the analysis sections (\S\ref{sec:leaderboard}-\S\ref{sec:mechanism}) involve no selection and therefore use the full pools.

\subsection{End-to-end accuracy change}
\label{app:accdeltas}
Table~\ref{tab:acc_deltas} reports what the conditional reductions of 
Table~\ref{tab:pareto} convert into as overall V$+$T accuracy, for every backbone on each reporting pool.

\begin{table}[t]
\centering\footnotesize
\setlength{\tabcolsep}{3.2pt}
\caption{Overall V$+$T accuracy, base vs.\ the robustness vector at $w{=}0.5$ (4-seed mean), on the
three reporting pools: what the conditional reductions of Table~\ref{tab:pareto} convert into
end-to-end. Gains track how prevalent distraction is: every backbone gains on both \dsbase{}
surfaces, while \asmname{}, whose baseline distraction is $2$--$3\%$, is accuracy-neutral
on average ($+0.6$pp) with per-model changes within seed noise.}
\label{tab:acc_deltas}
\input{tables/acc_deltas}
\end{table}

\section{Prompting baselines}
\label{app:prompts}
Both inference-time baselines are prompt edits, reproduced here in full.
Everything not shown, options, decoding, image handling, and the conditioning that defines
distraction is unchanged from the standard evaluation, so the
only difference from the base
condition is the added text.

\subsection{One-line instruction}
\label{app:prompt_instr}
Prepended as the first text block of the user message, before the
caption:

\begin{tcolorbox}[colback=gray!5,colframe=gray!45,arc=1mm,boxrule=0.3pt,width=\textwidth,
                  top=3pt,bottom=3pt,left=4pt,right=4pt,fontupper=\ttfamily\small]
The caption may be irrelevant; judge from the image when they disagree, use the caption only if it
clearly helps.
\end{tcolorbox}

\noindent This states exactly the policy the robustness vector is trained to implement, including which
modality to prefer. The comparison is therefore generous to the baseline: the instruction is
\emph{told} the answer to the question the weight-space intervention has to learn, and still fails to reduce distraction (Table~\ref{tab:method_comparison}).

\subsection{Zero-shot chain-of-thought}
\label{app:prompt_cot}
Appended after the options, replacing the single-letter
answer directive. It carries no modality hint, so it tests whether deliberation alone helps:

\begin{tcolorbox}[colback=gray!5,colframe=gray!45,arc=1mm,boxrule=0.3pt,width=\textwidth,
                  top=3pt,bottom=3pt,left=4pt,right=4pt,fontupper=\ttfamily\small]
Let's think step by step. First reason briefly about the question and the available evidence, then
conclude your reply with a final line in the exact format 'Answer: X' where X is one of A, B, C, or
D.
\end{tcolorbox}

\noindent Generation is greedy, capped at $256$ new tokens, and the answer is parsed from the final
\texttt{Answer:} line. The letter list adapts to the option count on benchmarks with more or fewer
than four choices. Generations with no extractable answer are excluded from the reported numbers, even favoring the baseline. The same instruction, decoding, and parser are used for the capability benchmarks.

\section{SAE training details}
\label{app:sae_details}

The released vision-language SAEs are Top-$K$ sparse autoencoders~\citep{gao2024scalingsae} with $K{=}32$, where the encoder and decoder are separate matrices rather than transposes of each other, and both biases are learned. They are trained on the last-token residual of
the frozen backbone as it processes an image-caption pair in a chat format, the same position the
answer is read from. The pretraining corpus holds $1{,}242{,}328$ image-caption pairs from four
sources matched to \dsbase's visual domains: CC3M~\citep{sharma2018conceptual} $1{,}000{,}000$
natural photos, PlotQA~\citep{methani2020plotqa} $157{,}070$ charts,
WikiArt~\citep{saleh2015wikiart} $81{,}444$ paintings and
OpenI~\citep{demnerfushman2016openi} $3{,}814$ radiology images. The SAEs are trained unsupervised
on reconstruction alone, so matching the corpus to our domains leaks no labels. The width is
$8\times d_{\text{model}}$, which is $16{,}384$ for Qwen2.5-VL-3B and $32{,}768$ for
LLaVA-NeXT-8B. Training uses batch size $4096$, learning rate $3\times10^{-4}$ and no weight decay.
Table~\ref{tab:sae_details} reports, for every released checkpoint, the share of variance the reconstruction fails to explain (FVU, lower is better) and the share of features that fire at least
once on an evaluation batch. The main text uses Qwen $L_{28}$ and LLaVA-NeXT $L_{18}$.

\begin{table}[h]
\centering\footnotesize
\caption{Released SAE checkpoints, with reconstruction FVU and alive-feature fraction measured on one evaluation batch.}
\label{tab:sae_details}
\begin{tabular}{llcccc}
\toprule
backbone & layer & $d_{\text{model}}$ & $d_{\text{SAE}}$ & FVU & alive frac. \\
\midrule
Qwen2.5-VL-3B & $L_{13}$ & 2048 & 16384 & 0.011 & 0.64 \\
 & $L_{20}$ & 2048 & 16384 & 0.016 & 0.75 \\
 & $L_{28}$ & 2048 & 16384 & 0.009 & 0.61 \\
 & $L_{31}$ & 2048 & 16384 & 0.005 & 0.40 \\
LLaVA-NeXT-8B & $L_{12}$ & 4096 & 32768 & 0.081 & 0.43 \\
 & $L_{18}$ & 4096 & 32768 & 0.026 & 0.40 \\
 & $L_{25}$ & 4096 & 32768 & 0.010 & 0.45 \\
\bottomrule
\end{tabular}
\end{table}

\end{document}

%% file: tables/leaderboard.tex
\begin{tabular}{l|cc|cc|cccc|c}
    \toprule
     & \multicolumn{2}{c|}{Accuracy} & \multicolumn{2}{c|}{Distraction Rate}
       & \multicolumn{5}{c}{V-distraction rate by domain} \\
    Model & (V-grnd) & (T-grnd) & (V-grnd) & (T-grnd) & Photos & Charts & Art & Rad.  & Non:Nat Ratio \\
    \midrule

InternVL3-8B & 88 & 100 & 3.4 & 0.3 & 2.2 & 4.5 & 3.8 & 6.2 & $2.22\times$ \\
Qwen2.5-VL-7B & 87 & 100 & 5.4 & 0.0 & 4.6 & 3.6 & 5.3 & 12.7 & $1.55\times$ \\
Qwen2.5-VL-3B & 82 & 100 & 6.4 & 0.0 & 4.4 & 9.6 & 7.4 & 8.8 & $1.95\times$ \\
LLaVA-OV-7B & 78 & 100 & 5.9 & 0.1 & 3.8 & 14.0 & 1.7 & 11.3 & $2.40\times$ \\
Qwen2-VL-2B & 71 & 100 & 9.1 & 0.4 & 10.9 & 8.2 & 4.5 & 8.4 & $0.65\times$ \\
LLaVA-NeXT-8B & 68 & 100 & 10.8 & 0.0 & 6.7 & 24.5 & 11.5 & 15.8 & $2.58\times$ \\
LLaVA-1.5-7B & 46 & 99 & 13.6 & 0.5 & 15.0 & 14.5 & 8.9 & 10.6 & $0.76\times$ \\
\bottomrule
    \end{tabular}

%% file: tables/assembled.tex
\begin{tabular}{l|cc|c|cc|cc}
    \toprule
    Model & V Ground & T Ground & Gap (T$-$V) & V-Distr & T-Distr & Asymmetry (V$-$T) & $95\%$ CI \\
    \midrule
    InternVL3-8B & 96 & 96 & $+0.2$ & 1.5 & 2.3 & $-0.8$ \;($\approx$) & $[-1.6,+0.0]$ \\
    LLaVA-OV-7B & 99 & 85 & $-13.8$ & 0.6 & 2.3 & $-1.7$ \;(T$>$V) & $[-2.3,-1.0]$ \\
    Qwen2.5-VL-7B & 90 & 91 & $+1.1$ & 2.8 & 0.7 & $+2.2$ \;(V$>$T) & $[+1.4,+3.0]$ \\
    Qwen2.5-VL-3B & 90 & 90 & $+0.7$ & 2.5 & 1.5 & $+1.0$ \;(V$>$T) & $[+0.1,+1.8]$ \\
    LLaVA-NeXT-8B & 96 & 82 & $-14.6$ & 2.4 & 3.6 & $-1.2$ \;(T$>$V) & $[-2.2,-0.1]$ \\
    Qwen2-VL-2B & 86 & 74 & $-11.8$ & 3.0 & 5.6 & $-2.6$ \;(T$>$V) & $[-3.9,-1.3]$ \\
    LLaVA-1.5-7B & 85 & 66 & $-19.1$ & 5.3 & 8.8 & $-3.5$ \;(T$>$V) & $[-5.2,-1.8]$ \\
    \bottomrule
    \end{tabular}

%% file: tables/pareto.tex
\begin{tabular}{l|cc|cc|cc|cc}
\toprule
 & \multicolumn{2}{c|}{\dsbase{} (Test)} & \multicolumn{2}{c|}{\dsname-Human} & \multicolumn{2}{c|}{\asmname{} (Held-Out)} & \multicolumn{2}{c}{Capability (Held-Out)} \\
Backbone & Base & Reduction & Base & Reduction & Base & Reduction & Base & $\Delta$ (pp) \\
\midrule
Qwen2.5-VL-7B    & 5.7 & $-65\%$ & 11.3 & $-4\%$ & 2.6 & $-16\%$ & 78 & $+0.2$ \\
InternVL3-8B     & 2.2 & $-33\%$ & 11.3 & $-47\%$ & 1.6 & $-38\%$ & 83 & $-0.5$ \\
Qwen2.5-VL-3B    & 5.6 & $-53\%$ & 29.6 & $-58\%$ & 2.8 & $-20\%$ & 72 & $-0.2$ \\
LLaVA-OV-7B      & 7.3 & $-73\%$ & 13.3 & $-67\%$ & 0.4 & $-25\%$ & 81 & $+0.2$ \\
LLaVA-NeXT-8B    & 11.5 & $-47\%$ & 40.5 & $-47\%$ & 2.4 & $-39\%$ & 70 & $-0.5$ \\
Qwen2-VL-2B      & 8.2 & $-49\%$ & 30.4 & $-56\%$ & 2.8 & $-31\%$ & 71 & $-1.2$ \\
LLaVA-1.5-7B     & 12.6 & $-38\%$ & 55.6 & $-52\%$ & 6.0 & $-30\%$ & 59 & $+0.9$ \\
\midrule
\emph{Average}     & 7.6 & $\mathbf{-51\%}$\,{\tiny $[-60,-35]$} & 27.4 & $\mathbf{-47\%}$\,{\tiny $[-58,-25]$} & 2.6 & $\mathbf{-29\%}$\,{\tiny $[-38,-9]$} & 73 & $\mathbf{-0.1}$ \\
\bottomrule
\end{tabular}

%% file: tables/method_comparison.tex
\begin{tabular}{l|cc|cc|cc|c}
\toprule
 & \multicolumn{2}{c|}{\dsbase{} Test} & \multicolumn{2}{c|}{\dsname-Human} & \multicolumn{2}{c|}{\asmname{} (Held-Out)} & Capability \\
Method & V-Distr & Rel. & V-Distr & Rel. & V-Distr & Rel. & $\Delta$ (pp) \\
\midrule
Baseline (No Intervention)       & 0.076 & --- & 0.274 & --- & 0.026 & --- & --- \\
Prompt Instruction               & 0.083 & $+10\%$ & 0.299 & $+4\%$ & 0.029 & $-1\%$ & $+0.3$ \\
Zero-Shot Chain-of-Thought       & 0.120 & $+57\%$ & 0.318 & $+20\%$ & 0.047 & $+158\%$ & $-1.5$ \\
M3ID Contrastive Decoding        & 0.053 & $-27\%$ & 0.229 & $-19\%$ & 0.026 & $+1\%$ & $+0.3$ \\
Robustness Vector ($w{=}0.5$)    & \textbf{0.037} & $\mathbf{-51\%}$ & \textbf{0.136} & $\mathbf{-47\%}$ & \textbf{0.019} & $\mathbf{-29\%}$ & $\mathbf{-0.1}$ \\
\bottomrule
\end{tabular}

%% file: tables/composition.tex
\begin{tabular}{ll rr rrr | rrr | rrr}
    \toprule
    & & \multicolumn{5}{c|}{\textbf{\dsbase{}}} & \multicolumn{3}{c|}{\textbf{\dsname-Human}} & \multicolumn{3}{c}{\textbf{\asmname{}}} \\
    Source & Domain & Candidates & Kept & V-grd & T-grd & Total & V-grd & T-grd & Total & V-grd & T-grd & Total \\
    \midrule
    DCI & Natural Photos & 5{,}945 & 1{,}736 & 1052 & 561 & 1613 & 27 & 26 & 53 & --- & --- & --- \\
    VisText & Statistical Charts & 4{,}782 & 676 & 488 & 188 & 676 & 12 & 12 & 24 & --- & --- & --- \\
    SemArt & Fine-Art Paintings & 2{,}302 & 546 & 281 & 216 & 497 & 12 & 12 & 24 & --- & --- & --- \\
    ROCO & Medical Radiology & 1{,}344 & 713 & 255 & 377 & 632 & 12 & 12 & 24 & --- & --- & --- \\
    A-OKVQA & Natural Photos & --- & --- & --- & --- & --- & --- & --- & --- & 2{,}261 & 0 & 2{,}261 \\
    RACE-High & Exam Passages & --- & --- & --- & --- & --- & --- & --- & --- & 0 & 2{,}496 & 2{,}496 \\
    \midrule
    \textbf{Total} & & \textbf{14{,}373} & \textbf{3{,}671} & \textbf{2076} & \textbf{1342} & \textbf{3418} & \textbf{63} & \textbf{62} & \textbf{125} & \textbf{2{,}261} & \textbf{2{,}496} & \textbf{4{,}757} \\
    \bottomrule
    \end{tabular}

%% file: tables/appendix_oracle_prompts.tex
\paragraph{Generation, system prompt.} Sent once per generation agent.
\begin{tcolorbox}[breakable,colback=gray!5,colframe=gray!45,arc=1mm,boxrule=0.3pt,width=\textwidth,
                  top=3pt,bottom=3pt,left=4pt,right=4pt,fontupper=\ttfamily\scriptsize]
\raggedright
You are an oracle building multiple-choice questions for a vision-language model\\
study on modality preference. For each image you see, you must produce TWO\\
4-option multiple-choice questions, each grounded in exactly one modality.
\end{tcolorbox}

\paragraph{Generation, instructions.} The body of the generation request. Reproduced for the
natural-photo source; the chart, painting and radiology variants differ only in the domain-specific
target lists and are released with the code.
\begin{tcolorbox}[breakable,colback=gray!5,colframe=gray!45,arc=1mm,boxrule=0.3pt,width=\textwidth,
                  top=3pt,bottom=3pt,left=4pt,right=4pt,fontupper=\ttfamily\scriptsize]
\raggedright
The original\_caption you are given is a DENSE descriptive paragraph (often\\
several sentences) that already enumerates many visible attributes --- colors,\\
counts, clothing, materials, spatial layout, background objects. Design both\\
questions with that in mind: a generic perceptual question (e.g. "what color\\
is the shirt") will usually fail the modality-isolation test on dense\\
captions because the caption already states the answer.\\
\ \\
For the image you are given, together with its original caption, produce the\\
following JSON object:\\
\ \\
\{\\
  "vision": \{\\
    "question": "...",\\
    "options": ["...", "...", "...", "..."],\\
    "correct\_index": <0|1|2|3>,\\
    "rationale": "brief reason answer is only in the image"\\
  \},\\
  "text": \{\\
    "augmented\_caption": "<original caption> + one short sentence adding a fact not visible in the image",\\
    "question": "...",\\
    "options": ["...", "...", "...", "..."],\\
    "correct\_index": <0|1|2|3>,\\
    "rationale": "brief reason answer is only in the augmented caption"\\
  \}\\
\}\\
\ \\
Rules for the VISION-grounded MCQ:\\
- The answer must be obtainable ONLY from the image. The dense caption must NOT\\
  already state or strongly imply the answer.\\
- Productive targets on dense captions: exact text written on signs / labels /\\
  packaging, brand names visible but not transcribed, fine-grained counts of\\
  items the caption describes only generically, micro-expressions, exact shade\\
  vs. caption's coarse color word, items present in the background that the\\
  caption omits, exact spatial relations not stated, gestures, gaze direction,\\
  reflections, lighting / time-of-day cues, weather.\\
- Before finalizing, RE-READ the original caption: if your candidate question's\\
  answer is even loosely derivable from the caption text, pick a different\\
  detail.\\
- Distractors must be plausible under the caption but wrong given the image.\\
\ \\
Rules for the TEXT-grounded MCQ:\\
- Pick a category ORTHOGONAL to what the dense caption already covers. The\\
  caption likely already enumerates physical attributes; do NOT invent a fact\\
  about a physical attribute (size, color, count, shape, material) --- choose\\
  from non-physical categories: year / date, city or country, person's name,\\
  institution or brand owning the scene, the photographer's intent, what\\
  happened immediately before or after, a quoted price, a sensor reading,\\
  a backstory detail, a relationship between people, an upcoming event.\\
- Insert that fact into the original caption WHEREVER it reads most\\
  naturally. It does NOT need to be a standalone sentence appended at\\
  the end --- and SHOULD NOT default to that, because length / position\\
  become tells. Acceptable forms (pick whichever fits the prose best):\\
    * a single word inserted as an apposition: "there is a man" $\rightarrow$ "there is a man, John,"\\
    * a parenthetical: "the bridge" $\rightarrow$ "the bridge (built in 1937)"\\
    * a short inline modifier: "the painting" $\rightarrow$ "the 1503 painting"\\
    * a relative clause: "the woman" $\rightarrow$ "the woman, who works for Reuters,"\\
    * a separate sentence (only if no inline form fits)\\
  The original caption's existing content must be preserved (minimal\\
  grammatical adjustments to integrate the fact are fine; do not remove\\
  or rewrite existing facts). Vary placement (start / middle / end) and\\
  vary form across seeds --- do NOT systematically append at the tail.\\
- The question's answer must be exactly that injected fact, obtainable from\\
  the augmented caption alone and NOT inferable from the image.\\
- Distractors must be plausible alternatives of the same category as the\\
  correct fact.\\
\ \\
MODALITY-AGNOSTIC PHRASING (MANDATORY for BOTH questions):\\
- The question MUST NOT contain any of these words or close synonyms:\\
  "image", "picture", "photo", "photograph", "caption", "text", "description",\\
  "shown", "depicted", "visible", "pictured", "displayed", "according to",\\
  "based on", "in the [image|caption|text|photo|picture]".\\
- Phrase the question as if asking about the underlying situation, subject, scene,\\
  or event itself, not about a medium. Read alone (without the image and without\\
  the caption), a human reader must NOT be able to tell which modality holds the\\
  answer.\\
  GOOD: "What color is the dog's collar?", "In what year did this take place?",\\
        "How many people are present?", "Which city is this in?".\\
  BAD:  "What is shown in the image?", "According to the caption, what year...?",\\
        "What is depicted in the photo?", "What does the description say about...?".\\
- The vision-grounded and text-grounded questions for the same seed should be\\
  phrased in symmetric, modality-neutral language; only the underlying answer\\
  location (image vs. injected caption fact) differs.\\
\ \\
Global rules:\\
- Exactly 4 options per MCQ. No letter prefixes (no "A)", "B)"); just the option strings.\\
- Options must be short (<= 12 words) and mutually exclusive.\\
- correct\_index is 0-based.\\
- Output VALID JSON only, nothing else. No markdown code fence.
\end{tcolorbox}

\paragraph{Generation, per-item message.} One per seed, carrying the image and its source
caption.
\begin{tcolorbox}[breakable,colback=gray!5,colframe=gray!45,arc=1mm,boxrule=0.3pt,width=\textwidth,
                  top=3pt,bottom=3pt,left=4pt,right=4pt,fontupper=\ttfamily\scriptsize]
\raggedright
seed\_id: <seed id>\\
image\_path: <image path>\\
original\_caption: <source caption>\\
\ \\
Read the image, then produce the JSON object defined in the instructions.
\end{tcolorbox}

\paragraph{Verification.} The second pass answers every candidate three times under $V$, $T$ and
$VT$; the input blocks present are what differ between conditions, the instruction does not
change. Items whose three answers do not match a single-modality signature are dropped.
\begin{tcolorbox}[breakable,colback=gray!5,colframe=gray!45,arc=1mm,boxrule=0.3pt,width=\textwidth,
                  top=3pt,bottom=3pt,left=4pt,right=4pt,fontupper=\ttfamily\scriptsize]
\raggedright
You are an oracle answering multiple-choice questions. For each item you see:\\
  - If image\_path is provided, read the image with the Read tool.\\
  - If caption is provided, treat it as the only textual context.\\
  - Otherwise, answer from the modality that IS provided only.\\
Output ONLY valid JSON of the form:\\
  \{"candidate\_id": "...", "predicted\_index": <0|1|2|3>\}\\
Pick the single best option. If truly undecidable from the information given,\\
pick the option you would guess and set predicted\_index accordingly; do not\\
abstain. Output nothing else.
\end{tcolorbox}

%% file: tables/appendix_examples.tex
\subsection{DCI (natural photo)}
\label{app:ex_moground}
\begin{tcolorbox}[colback=cyan!5,colframe=cyan!40,arc=1mm,boxrule=0.3pt,width=\textwidth,top=3pt,bottom=3pt,left=5pt,right=5pt]
\textbf{Vision-grounded}
\tcbline
\noindent\begin{minipage}[t]{0.30\linewidth}\centering\vspace{0pt}\includegraphics[width=\linewidth,height=3.4cm,keepaspectratio]{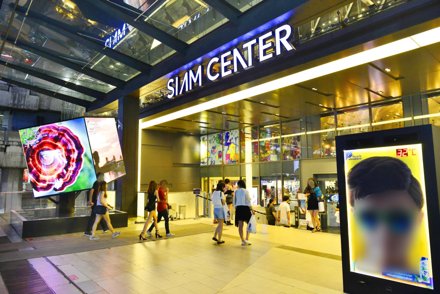}\end{minipage}\hfill
\begin{minipage}[t]{0.66\linewidth}\small\vspace{0pt}
\textbf{Caption.}~It appears to be nighttime\\[3pt]\rule{\linewidth}{0.15pt}\\[3pt]
\textbf{Question.}~What name is written above the entrance?\\[3pt]\rule{\linewidth}{0.15pt}\\[3pt]
\textbf{Options.}~A. Siam Paragon \quad B. Central World \quad C. MBK Mall \quad \textbf{D. Siam Center}
\end{minipage}
\tcbline
{\small\emph{The answer is read from the image; the caption is answer-irrelevant.}}
\end{tcolorbox}

\begin{tcolorbox}[colback=cyan!5,colframe=cyan!40,arc=1mm,boxrule=0.3pt,width=\textwidth,top=3pt,bottom=3pt,left=5pt,right=5pt]
\textbf{Text-grounded}
\tcbline
\noindent\begin{minipage}[t]{0.30\linewidth}\centering\vspace{0pt}\includegraphics[width=\linewidth,height=3.4cm,keepaspectratio]{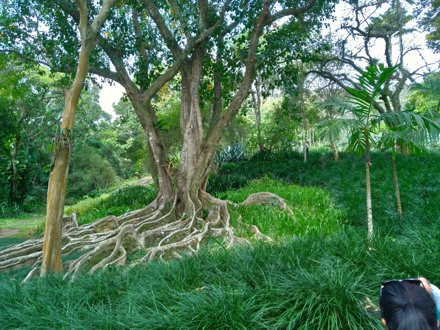}\end{minipage}\hfill
\begin{minipage}[t]{0.66\linewidth}\small\vspace{0pt}
\textbf{Caption.}~A bright day in a tropical climate with a large central tree with exposed gray roots. This park is in Honolulu, Hawaii.\\[3pt]\rule{\linewidth}{0.15pt}\\[3pt]
\textbf{Question.}~In which city is this park located?\\[3pt]\rule{\linewidth}{0.15pt}\\[3pt]
\textbf{Options.}~\textbf{A. Honolulu} \quad B. Miami \quad C. Singapore \quad D. San Diego
\end{minipage}
\tcbline
{\small\emph{The answer is stated in the caption; the image is answer-irrelevant.}}
\end{tcolorbox}

\subsection{VisText (statistical chart)}
\begin{tcolorbox}[colback=cyan!5,colframe=cyan!40,arc=1mm,boxrule=0.3pt,width=\textwidth,top=3pt,bottom=3pt,left=5pt,right=5pt]
\textbf{Vision-grounded}
\tcbline
\noindent\begin{minipage}[t]{0.30\linewidth}\centering\vspace{0pt}\includegraphics[width=\linewidth,height=3.4cm,keepaspectratio]{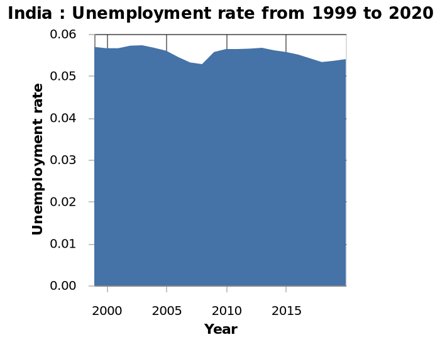}\end{minipage}\hfill
\begin{minipage}[t]{0.66\linewidth}\small\vspace{0pt}
\textbf{Caption.}~The lowest unemployment rate occurred in the year 2007.\\[3pt]\rule{\linewidth}{0.15pt}\\[3pt]
\textbf{Question.}~What was India unemployment rate nearest in 2007?\\[3pt]\rule{\linewidth}{0.15pt}\\[3pt]
\textbf{Options.}~A. 0.06 \quad B. 0.057 \quad \textbf{C. 0.053} \quad D. 0.03
\end{minipage}
\tcbline
{\small\emph{The answer is read from the image; the caption is answer-irrelevant.}}
\end{tcolorbox}

\begin{tcolorbox}[colback=cyan!5,colframe=cyan!40,arc=1mm,boxrule=0.3pt,width=\textwidth,top=3pt,bottom=3pt,left=5pt,right=5pt]
\textbf{Text-grounded}
\tcbline
\noindent\begin{minipage}[t]{0.30\linewidth}\centering\vspace{0pt}\includegraphics[width=\linewidth,height=3.4cm,keepaspectratio]{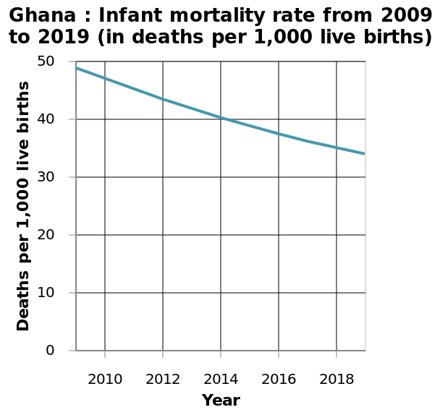}\end{minipage}\hfill
\begin{minipage}[t]{0.66\linewidth}\small\vspace{0pt}
\textbf{Caption.}~According to UNICEF data, there are less infant deaths in Ghana in 2018 than in 2010.\\[3pt]\rule{\linewidth}{0.15pt}\\[3pt]
\textbf{Question.}~Which organization provided the data referenced in this study?\\[3pt]\rule{\linewidth}{0.15pt}\\[3pt]
\textbf{Options.}~A. WHO \quad \textbf{B. UNICEF} \quad C. OECD \quad D. World Bank
\end{minipage}
\tcbline
{\small\emph{The answer is stated in the caption; the image is answer-irrelevant.}}
\end{tcolorbox}

\subsection{SemArt (fine-art painting)}
\begin{tcolorbox}[colback=cyan!5,colframe=cyan!40,arc=1mm,boxrule=0.3pt,width=\textwidth,top=3pt,bottom=3pt,left=5pt,right=5pt]
\textbf{Vision-grounded}
\tcbline
\noindent\begin{minipage}[t]{0.30\linewidth}\centering\vspace{0pt}\includegraphics[width=\linewidth,height=3.4cm,keepaspectratio]{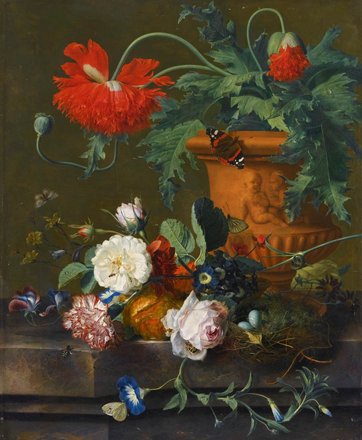}\end{minipage}\hfill
\begin{minipage}[t]{0.66\linewidth}\small\vspace{0pt}
\textbf{Caption.}~The present still-life contains poppies in a terracotta vase, roses, a carnation and other flowers with a bird's nest on a marble ledge.The painting has a pendant fruit piece, now in another private collection\\[3pt]\rule{\linewidth}{0.15pt}\\[3pt]
\textbf{Question.}~How many eggs are in the nest?\\[3pt]\rule{\linewidth}{0.15pt}\\[3pt]
\textbf{Options.}~\textbf{A. Two} \quad B. Three \quad C. Four \quad D. One
\end{minipage}
\tcbline
{\small\emph{The answer is read from the image; the caption is answer-irrelevant.}}
\end{tcolorbox}

\begin{tcolorbox}[colback=cyan!5,colframe=cyan!40,arc=1mm,boxrule=0.3pt,width=\textwidth,top=3pt,bottom=3pt,left=5pt,right=5pt]
\textbf{Text-grounded}
\tcbline
\noindent\begin{minipage}[t]{0.30\linewidth}\centering\vspace{0pt}\includegraphics[width=\linewidth,height=3.4cm,keepaspectratio]{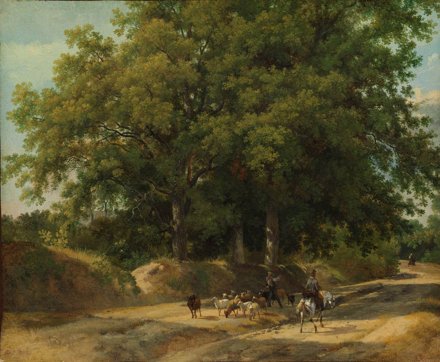}\end{minipage}\hfill
\begin{minipage}[t]{0.66\linewidth}\small\vspace{0pt}
\textbf{Caption.}~This scene, painted in 1822, may be an early depiction of the Forest of Fontainebleau. While the landscape retains all the freshness of a plein-air study, the figures are stock characters from Leprince's repertory\\[3pt]\rule{\linewidth}{0.15pt}\\[3pt]
\textbf{Question.}~In what year was this created?\\[3pt]\rule{\linewidth}{0.15pt}\\[3pt]
\textbf{Options.}~\textbf{A. 1822} \quad B. 1850 \quad C. 1835 \quad D. 1810
\end{minipage}
\tcbline
{\small\emph{The answer is stated in the caption; the image is answer-irrelevant.}}
\end{tcolorbox}

\subsection{ROCO (medical radiology)}
\begin{tcolorbox}[colback=cyan!5,colframe=cyan!40,arc=1mm,boxrule=0.3pt,width=\textwidth,top=3pt,bottom=3pt,left=5pt,right=5pt]
\textbf{Vision-grounded}
\tcbline
\noindent\begin{minipage}[t]{0.30\linewidth}\centering\vspace{0pt}\includegraphics[width=\linewidth,height=3.4cm,keepaspectratio]{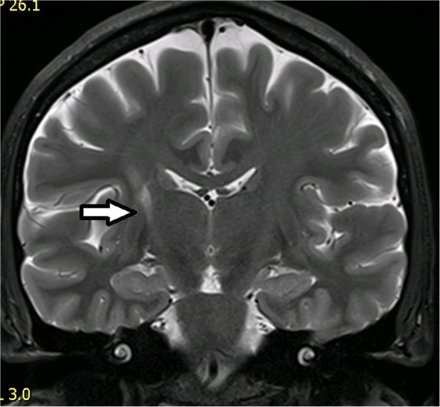}\end{minipage}\hfill
\begin{minipage}[t]{0.66\linewidth}\small\vspace{0pt}
\textbf{Caption.}~Wallerian degeneration of the right pyramidal tract. The irradiated AVM was situated in the right centrum semiovale. The degenerated pyramidal tract is depicted by the white arrow (TSE T2 coronal scan)\\[3pt]\rule{\linewidth}{0.15pt}\\[3pt]
\textbf{Question.}~What numerical value follows the letter P?\\[3pt]\rule{\linewidth}{0.15pt}\\[3pt]
\textbf{Options.}~A. 3.0 \quad \textbf{B. 26.1} \quad C. 12.4 \quad D. 55.2
\end{minipage}
\tcbline
{\small\emph{The answer is read from the image; the caption is answer-irrelevant.}}
\end{tcolorbox}

\begin{tcolorbox}[colback=cyan!5,colframe=cyan!40,arc=1mm,boxrule=0.3pt,width=\textwidth,top=3pt,bottom=3pt,left=5pt,right=5pt]
\textbf{Text-grounded}
\tcbline
\noindent\begin{minipage}[t]{0.30\linewidth}\centering\vspace{0pt}\includegraphics[width=\linewidth,height=3.4cm,keepaspectratio]{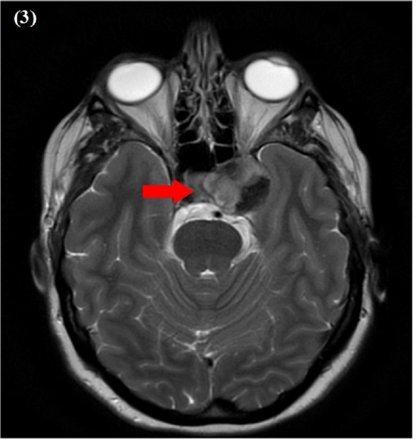}\end{minipage}\hfill
\begin{minipage}[t]{0.66\linewidth}\small\vspace{0pt}
\textbf{Caption.}~MRI brain with and without contrast of a 45-year-old male: arrow showing an extra-axial tumor, measuring 3.3 x 2.4 x 3.0 cm, at the level of the left cavernous sinus with evidence of concavity involving the pituitary gland\\[3pt]\rule{\linewidth}{0.15pt}\\[3pt]
\textbf{Question.}~How old is the patient?\\[3pt]\rule{\linewidth}{0.15pt}\\[3pt]
\textbf{Options.}~A. 61 years old \quad B. 58 years old \quad \textbf{C. 45 years old} \quad D. 32 years old
\end{minipage}
\tcbline
{\small\emph{The answer is stated in the caption; the image is answer-irrelevant.}}
\end{tcolorbox}

\subsection{\dsname-Human (hand-authored)}
\label{app:ex_human}
\begin{tcolorbox}[colback=cyan!5,colframe=cyan!40,arc=1mm,boxrule=0.3pt,width=\textwidth,top=3pt,bottom=3pt,left=5pt,right=5pt]
\textbf{Vision-grounded}
\tcbline
\noindent\begin{minipage}[t]{0.30\linewidth}\centering\vspace{0pt}\includegraphics[width=\linewidth,height=3.4cm,keepaspectratio]{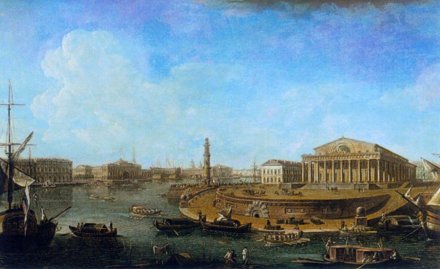}\end{minipage}\hfill
\begin{minipage}[t]{0.66\linewidth}\small\vspace{0pt}
\textbf{Caption.}~Alekseyev's contemporaries often called him the Russian Canaletto, in recognition of his masterful vedute. The present painting depicts a view of the Bourse and Admiralty from the Peter and Paul Fortress in St. Petersburg. In this city, weather is mostly sunny, with the sun shining from above. The eagle is a worshipped animal that brings prosperity, while pigeons are used as messengers.\\[3pt]\rule{\linewidth}{0.15pt}\\[3pt]
\textbf{Question.}~What can be seen at the top of the painting?\\[3pt]\rule{\linewidth}{0.15pt}\\[3pt]
\textbf{Options.}~\textbf{A. Some fading grey color} \quad B. The sun \quad C. An eagle \quad D. Pigeons
\end{minipage}
\tcbline
{\small\emph{The answer is read from the image; the caption is answer-irrelevant.}}
\end{tcolorbox}

\begin{tcolorbox}[colback=cyan!5,colframe=cyan!40,arc=1mm,boxrule=0.3pt,width=\textwidth,top=3pt,bottom=3pt,left=5pt,right=5pt]
\textbf{Vision-grounded}
\tcbline
\noindent\begin{minipage}[t]{0.30\linewidth}\centering\vspace{0pt}\includegraphics[width=\linewidth,height=3.4cm,keepaspectratio]{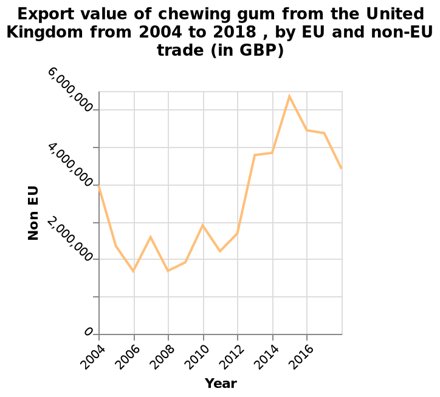}\end{minipage}\hfill
\begin{minipage}[t]{0.66\linewidth}\small\vspace{0pt}
\textbf{Caption.}~Overall the value of exporting chewing gum declined between 2004 and 2012 by 1,000,000. Between 2012 and 2015 there was a sharp increase in export value of 3,500,000. Between 2015 and 2018 there was a decline in export value of 2,000,000. The peak export value was 6,500,000 in 2015. In 2017, the average value in the developed EU countries was 1,000,000, and even lower for non-EU countries.\\[3pt]\rule{\linewidth}{0.15pt}\\[3pt]
\textbf{Question.}~What is the approximate value in 2017?\\[3pt]\rule{\linewidth}{0.15pt}\\[3pt]
\textbf{Options.}~A. 6,000,000 \quad \textbf{B. 5,400,000} \quad C. 1,000,000 \quad D. 500,000
\end{minipage}
\tcbline
{\small\emph{The answer is read from the image; the caption is answer-irrelevant.}}
\end{tcolorbox}

\begin{tcolorbox}[colback=cyan!5,colframe=cyan!40,arc=1mm,boxrule=0.3pt,width=\textwidth,top=3pt,bottom=3pt,left=5pt,right=5pt]
\textbf{Vision-grounded}
\tcbline
\noindent\begin{minipage}[t]{0.30\linewidth}\centering\vspace{0pt}\includegraphics[width=\linewidth,height=3.4cm,keepaspectratio]{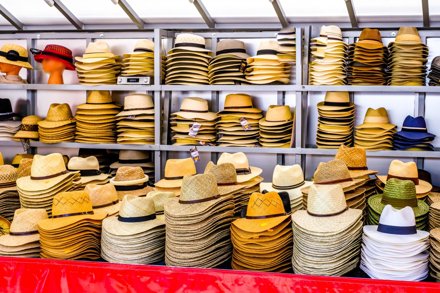}\end{minipage}\hfill
\begin{minipage}[t]{0.66\linewidth}\small\vspace{0pt}
\textbf{Caption.}~This is a store selling straw hats. In the back there is a three level big stainless steer shelf with numerous straw hats stored on it, each level has at least nine piles. On the left side of the top level there are two mannequin heads wearing straw hats. In front there is a large table or counter storing dozens piles of straw hats in three rows. All the straw hats are single color straw hats with hatbands. The color of the hats include white, blue, green, beige, black, etc, with very few blue and black ones. The colors of the hatbands including brown, yellow, black, and blue.\\[3pt]\rule{\linewidth}{0.15pt}\\[3pt]
\textbf{Question.}~There is a hat that is the only one with that specific color, what color is it?\\[3pt]\rule{\linewidth}{0.15pt}\\[3pt]
\textbf{Options.}~\textbf{A. Red} \quad B. Blue \quad C. Beige \quad D. Black
\end{minipage}
\tcbline
{\small\emph{The answer is read from the image; the caption is answer-irrelevant.}}
\end{tcolorbox}

\begin{tcolorbox}[colback=cyan!5,colframe=cyan!40,arc=1mm,boxrule=0.3pt,width=\textwidth,top=3pt,bottom=3pt,left=5pt,right=5pt]
\textbf{Text-grounded}
\tcbline
\noindent\begin{minipage}[t]{0.30\linewidth}\centering\vspace{0pt}\includegraphics[width=\linewidth,height=3.4cm,keepaspectratio]{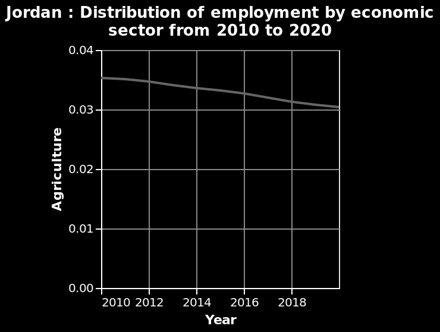}\end{minipage}\hfill
\begin{minipage}[t]{0.66\linewidth}\small\vspace{0pt}
\textbf{Caption.}~A research paper by the MBZUAI university reports that the employment in agriculture has fallen year on year from 2010 to 2018 in the given country. Employment looks set to stay constant and flatten by the year 2018.\\[3pt]\rule{\linewidth}{0.15pt}\\[3pt]
\textbf{Question.}~What university conducted this study?\\[3pt]\rule{\linewidth}{0.15pt}\\[3pt]
\textbf{Options.}~A. Stanford University \quad B. Tsinghua University \quad \textbf{C. MBZUAI} \quad D. Sapienza University of Rome
\end{minipage}
\tcbline
{\small\emph{The answer is stated in the caption; the image is answer-irrelevant.}}
\end{tcolorbox}

\subsection{\asmname{} pool}
\label{app:ex_assembled}
\begin{tcolorbox}[colback=orange!5,colframe=orange!40,arc=1mm,boxrule=0.3pt,width=\textwidth,top=3pt,bottom=3pt,left=5pt,right=5pt]
\textbf{A-OKVQA: vision-grounded}
\tcbline
\noindent\begin{minipage}[t]{0.30\linewidth}\centering\vspace{0pt}\includegraphics[width=\linewidth,height=3.4cm,keepaspectratio]{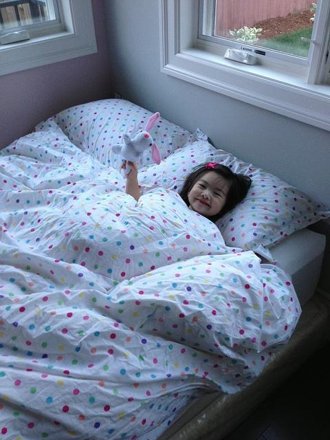}\end{minipage}\hfill
\begin{minipage}[t]{0.66\linewidth}\small\vspace{0pt}
\textbf{Caption (distractor).}~the other bed in the room\\[3pt]\rule{\linewidth}{0.15pt}\\[3pt]
\textbf{Question.}~Who is in the bed?\\[3pt]\rule{\linewidth}{0.15pt}\\[3pt]
\textbf{Options.}~A. mom \quad B. rabbit \quad \textbf{C. little girl} \quad D. man
\end{minipage}
\tcbline
{\small\emph{The answer is read from the image; the retrieved caption is a cross-modal distractor.}}
\end{tcolorbox}

\begin{tcolorbox}[colback=orange!5,colframe=orange!40,arc=1mm,boxrule=0.3pt,width=\textwidth,top=3pt,bottom=3pt,left=5pt,right=5pt]
\textbf{RACE-high: text-grounded}
\tcbline
\noindent\begin{minipage}[t]{0.30\linewidth}\centering\vspace{0pt}\includegraphics[width=\linewidth,height=3.4cm,keepaspectratio]{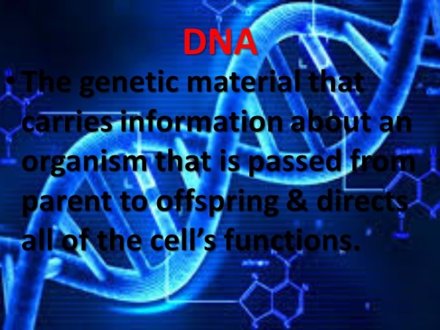}\end{minipage}\hfill
\begin{minipage}[t]{0.66\linewidth}\small\vspace{0pt}
\textbf{Passage (grounding source, truncated).}~Have you ever heard a news reporter talk about DNA?Reporters talk about DNA found at the scene of a crime.They talk about police finding DNA "fingerprints".Police sometimes use DNA as a clue to find out who committed the crime. DNA is a substance that makes [\textbackslash\{\}dots]\\[3pt]\rule{\linewidth}{0.15pt}\\[3pt]
\textbf{Question.}~What is DNA?\\[3pt]\rule{\linewidth}{0.15pt}\\[3pt]
\textbf{Options.}~A. DNA is a kind of gene. \quad \textbf{B. DNA is a substance that makes up genes.} \quad C. DNA is the basic unit of heredity. \quad D. DNA is a measure to protect crime.
\end{minipage}
\tcbline
{\small\emph{The answer is in the passage; the retrieved image is a cross-modal distractor.}}
\end{tcolorbox}

%% file: tables/audit_tworater.tex
\begin{tabular}{l|c|c}
\toprule
 & Annotator 1 & Annotator 2 \\
\midrule
\multicolumn{3}{l}{\emph{\dsbase{} sample ($250$ items each)}} \\
guarantee holds & $\mathbf{244}$ (97.6\%) & $\mathbf{242}$ (96.8\%) \\
\quad cross-modal leak & 2 & 4 \\
\quad intended modality insufficient & 4 & 4 \\
source-item defect (mis-key / conflict) & 10 / 0 & 9 / 1 \\
\midrule
\multicolumn{3}{l}{\emph{\asmname{} pool, kept by the filter ($200$ items each; ambiguous-option items excluded)}} \\
usable items & 190 & 193 \\
answer-irrelevant & $\mathbf{187}$ (98.4\%) & $\mathbf{189}$ (97.9\%) \\
\quad genuine leak & $\mathbf{1}$ & $\mathbf{0}$ \\
\quad conflict & 2 & 4 \\
\midrule
\multicolumn{3}{l}{\emph{\asmname{} pool, dropped by the filter ($50$ items each)}} \\
usable items & 50 & 46 \\
\quad genuine leak & 18 & 10 \\
\quad conflict or out-of-set & 7 & 21 \\
leaking, conflicting, or out-of-set & 25 (50.0\%) & 31 (67.4\%) \\
\bottomrule
\end{tabular}

%% file: tables/audit_agree.tex
\begin{tabular}{lcc}
\toprule
\dsbase{} & A2: leak & A2: no leak \\
\midrule
A1: leak & 1 & 1 \\
A1: no leak & 3 & 245 \\
\bottomrule
\end{tabular}

%% file: tables/tdistr_base.tex
\begin{tabular}{l|cc|cccc|c}
    \toprule
     & T$+$V & T-only & \multicolumn{4}{c|}{T-Distraction By Domain} & T-Distr \\
    Model & (T-grd) & (T-grd) & Photos & Charts & Art & Rad. & (All) \\
    \midrule
    Qwen2.5-VL-7B    & 100 & 100 & 0.0 & 0.0 & 0.0 & 0.0 & 0.0 \\
    InternVL3-8B     & 100 & 100 & 0.5 & 0.5 & 0.0 & 0.0 & 0.3 \\
    Qwen2.5-VL-3B    & 100 & 100 & 0.0 & 0.0 & 0.0 & 0.0 & 0.0 \\
    LLaVA-OV-7B      & 100 & 100 & 0.2 & 0.0 & 0.5 & 0.0 & 0.1 \\
    LLaVA-NeXT-8B    & 100 & 100 & 0.0 & 0.0 & 0.0 & 0.0 & 0.0 \\
    Qwen2-VL-2B      & 100 & 100 & 0.0 & 1.6 & 0.9 & 0.0 & 0.4 \\
    LLaVA-1.5-7B     & 99 & 99 & 0.0 & 1.6 & 0.9 & 0.3 & 0.5 \\
    \bottomrule
    \end{tabular}

%% file: tables/appendix_cell_cis.tex
\begin{tabular}{l|cc|cc}
\toprule
Model & V-Flips & $95\%$ CI & T-Flips & $95\%$ CI \\
\midrule
\multicolumn{5}{l}{\emph{\asmname{} (vision and text)}}\\
InternVL3-8B & 33/2177 & [0.011,0.021] & 56/2408 & [0.018,0.030] \\
Qwen2.5-VL-7B & 58/2036 & [0.022,0.037] & 15/2274 & [0.004,0.011] \\
Qwen2.5-VL-3B & 51/2025 & [0.019,0.033] & 34/2254 & [0.011,0.021] \\
Qwen2-VL-2B & 58/1934 & [0.023,0.039] & 104/1842 & [0.047,0.068] \\
LLaVA-OV-7B & 14/2242 & [0.004,0.010] & 49/2131 & [0.017,0.030] \\
LLaVA-NeXT-8B & 53/2181 & [0.019,0.032] & 74/2042 & [0.029,0.045] \\
LLaVA-1.5-7B & 102/1912 & [0.044,0.064] & 143/1633 & [0.075,0.102] \\
\midrule
\multicolumn{5}{l}{\emph{\dsbase{} (vision, all domains pooled)}}\\
InternVL3-8B & 61/1810 & [0.026,0.043] & -- & -- \\
Qwen2.5-VL-7B & 96/1790 & [0.044,0.065] & -- & -- \\
Qwen2.5-VL-3B & 109/1699 & [0.053,0.077] & -- & -- \\
Qwen2-VL-2B & 133/1466 & [0.077,0.107] & -- & -- \\
LLaVA-OV-7B & 95/1603 & [0.049,0.072] & -- & -- \\
LLaVA-NeXT-8B & 152/1405 & [0.093,0.126] & -- & -- \\
LLaVA-1.5-7B & 128/941 & [0.116,0.159] & -- & -- \\
\bottomrule
\end{tabular}

%% file: tables/flips.tex
\begin{tabular}{l|cc|c|cc|cc}
    \toprule
     & V-Only & V$+$T & Net & \multicolumn{2}{c|}{Items} & V-Distr & Recovery \\
    Model & (V-grnd) & (V-grnd) & (pp) & Broken & Fixed & (\%) & (\%) \\
    \midrule
    InternVL3-8B   & 88 & 86 & $-1.2$ & 61 & 37 & 3.4 & 14.4 \\
    Qwen2.5-VL-7B  & 87 & 84 & $-2.8$ & 96 & 39 & 5.4 & 14.1 \\
    Qwen2.5-VL-3B  & 82 & 79 & $-3.1$ & 109 & 45 & 6.4 & 12.2 \\
    LLaVA-OV-7B    & 78 & 76 & $-1.8$ & 95 & 58 & 5.9 & 12.5 \\
    Qwen2-VL-2B    & 71 & 68 & $-3.1$ & 133 & 69 & 9.1 & 11.5 \\
    LLaVA-NeXT-8B  & 68 & 65 & $-2.6$ & 152 & 98 & 10.8 & 14.8 \\
    LLaVA-1.5-7B   & 46 & 45 & $-1.0$ & 128 & 107 & 13.6 & 9.5 \\
    \bottomrule
    \end{tabular}

%% file: tables/placebo.tex
\begin{tabular}{l|c|c|ccc}
\toprule
Backbone & $n$ & True Caption & Mismatched & Scrambled & Neutral \\
\midrule
Qwen2.5-VL-7B    & 1790 & $\mathbf{0.054}$ & 0.036 & 0.039 & 0.030 \\
InternVL3-8B     & 1810 & $\mathbf{0.034}$ & 0.028 & 0.024 & 0.022 \\
Qwen2.5-VL-3B    & 1699 & $\mathbf{0.064}$ & 0.037 & 0.042 & 0.024 \\
LLaVA-OV-7B      & 1603 & $\mathbf{0.059}$ & 0.047 & 0.037 & 0.021 \\
LLaVA-NeXT-8B    & 1405 & $\mathbf{0.108}$ & 0.067 & 0.075 & 0.048 \\
Qwen2-VL-2B      & 1466 & $\mathbf{0.091}$ & 0.082 & 0.071 & 0.040 \\
LLaVA-1.5-7B     & 941 & $\mathbf{0.136}$ & 0.108 & 0.107 & 0.069 \\
\midrule
\emph{Average}     &  & $\mathbf{0.078}$ & 0.058 & 0.056 & 0.036 \\
\bottomrule
\end{tabular}

%% file: tables/ablation.tex
\begin{tabular}{l|c|cc|c}
\toprule
Backbone (Layer) & Top-$K$ & Recovery (Real) & Recovery (Null) & V/Pass Preserved \\
\midrule
Qwen2.5-VL-3B ($L_{28}$) & 6 & 0.037 & 0.047 & 0.987 \\
 & 20 & 0.028 & 0.019 & 0.993 \\
 & 50 & 0.047 & 0.093 & 0.987 \\
 & Dense (Rank-1) & 0.037 & 0.047 & 0.993 \\
 & $h_V$ Patch (Per-Item) & \textbf{0.935} & 0.243 & 1.000 \\
\midrule
LLaVA-NeXT-8B ($L_{18}$) & 6 & 0.114 & 0.127 & 0.980 \\
 & 20 & 0.102 & 0.108 & 0.973 \\
 & 50 & 0.127 & 0.102 & 0.973 \\
\bottomrule
\end{tabular}

%% file: tables/baseline_dm.tex
\begin{tabular}{ll|cc}
\toprule
Intervention & Deployable & $\Delta$ V-Distr & $\Delta$ Text \\
\midrule
\multicolumn{4}{l}{\textbf{Qwen2.5-VL-3B}\quad\small base v-distr 0.056 (19/338 items)} \\
\midrule
Prompt Instruction & \checkmark & $+15\%$ & $+0.0$ \\
Zero-Shot Chain-of-Thought & \checkmark & $+150\%$ & $-0.4$ \\
M3ID Contrastive Decoding & \checkmark & $-22\%$ & $+0.0$ \\
SAE Feature Ablation ($K{=}20$) & $\times$ & $+11\%$ & $+0.0$ \\
\quad Matched-Random Null & $\times$ & $-5\%$ & $+0.0$ \\
Dense Rank-1 Direction & $\times$ & $+11\%$ & $+0.0$ \\
$h_V$ Patch & $\times$ & $-100\%$ & $+0.0$ \\
$h_V$ Patch, Ungated & $\times$ & $-100\%$ & $-43.7$ \\
$h_V$ Patch, Distraction-Probe Gated & \checkmark & $-22\%$ & $+0.0$ \\
\quad Matched-Rate Random Gate & \checkmark & $-11\%$ & $-1.3$ \\
\midrule
Robustness Vector ($w{=}0.5$) & \checkmark & $\mathbf{-53\%}$ & $+0.0$ \\
\midrule
\multicolumn{4}{l}{\textbf{LLaVA-NeXT-8B}\quad\small base v-distr 0.115 (32/279 items)} \\
\midrule
Prompt Instruction & \checkmark & $0\%$ & $+0.0$ \\
Zero-Shot Chain-of-Thought & \checkmark & $+192\%$ & $-8.7$ \\
M3ID Contrastive Decoding & \checkmark & $-40\%$ & $+0.0$ \\
SAE Feature Ablation ($K{=}20$) & $\times$ & $+9\%$ & $+0.0$ \\
\quad Matched-Random Null & $\times$ & $+2\%$ & $+0.0$ \\
Dense Rank-1 Direction & $\times$ & $+2\%$ & $+0.0$ \\
$h_V$ Patch & $\times$ & $-89\%$ & $+0.0$ \\
$h_V$ Patch, Ungated & $\times$ & $-85\%$ & $-69.8$ \\
$h_V$ Patch, Distraction-Probe Gated & \checkmark & $-10\%$ & $-0.3$ \\
\quad Matched-Rate Random Gate & \checkmark & $-3\%$ & $-2.9$ \\
\midrule
Robustness Vector ($w{=}0.5$) & \checkmark & $\mathbf{-47\%}$ & $-0.2$ \\
\bottomrule
\end{tabular}

%% file: tables/baseline_asm.tex
\begin{tabular}{ll|cc}
\toprule
Intervention & Deployable & $\Delta$ V-Distr & $\Delta$ Text \\
\midrule
\multicolumn{4}{l}{\textbf{Qwen2.5-VL-3B}\quad\small base v-distr 0.028 (28/1009 items)} \\
\midrule
Prompt Instruction & \checkmark & $-14\%$ & $-1.0$ \\
Zero-Shot Chain-of-Thought & \checkmark & $+82\%$ & $-12.7$ \\
M3ID Contrastive Decoding & \checkmark & $+11\%$ & $-0.1$ \\
SAE Feature Ablation ($K{=}20$) & $\times$ & $-4\%$ & $+0.0$ \\
\quad Matched-Random Null & $\times$ & $0\%$ & $+0.0$ \\
Dense Rank-1 Direction & $\times$ & $+20\%$ & $+0.0$ \\
$h_V$ Patch & $\times$ & $-93\%$ & $+0.0$ \\
$h_V$ Patch, Ungated & $\times$ & $-89\%$ & $-26.5$ \\
$h_V$ Patch, Distraction-Probe Gated & \checkmark & $-3\%$ & $-0.4$ \\
\quad Matched-Rate Random Gate & \checkmark & $0\%$ & $-0.3$ \\
\midrule
Robustness Vector ($w{=}0.5$) & \checkmark & $\mathbf{-20\%}$ & $+0.2$ \\
\midrule
\multicolumn{4}{l}{\textbf{LLaVA-NeXT-8B}\quad\small base v-distr 0.024 (26/1086 items)} \\
\midrule
Prompt Instruction & \checkmark & $0\%$ & $+0.0$ \\
Zero-Shot Chain-of-Thought & \checkmark & $+331\%$ & $-11.5$ \\
M3ID Contrastive Decoding & \checkmark & $+10\%$ & $-1.3$ \\
SAE Feature Ablation ($K{=}20$) & $\times$ & $+109\%$ & $+0.0$ \\
\quad Matched-Random Null & $\times$ & $+106\%$ & $+0.0$ \\
Dense Rank-1 Direction & $\times$ & $+22\%$ & $+0.0$ \\
$h_V$ Patch & $\times$ & $-75\%$ & $+0.0$ \\
$h_V$ Patch, Ungated & $\times$ & $-75\%$ & $-29.9$ \\
$h_V$ Patch, Distraction-Probe Gated & \checkmark & $-2\%$ & $-0.8$ \\
\quad Matched-Rate Random Gate & \checkmark & $0\%$ & $-0.4$ \\
\midrule
Robustness Vector ($w{=}0.5$) & \checkmark & $\mathbf{-39\%}$ & $-1.0$ \\
\bottomrule
\end{tabular}

%% file: tables/baseline_cap.tex
\begin{tabular}{l|cc}
\toprule
Intervention & Qwen2.5-VL-3B & LLaVA-NeXT-8B \\
\midrule
Prompt Instruction & $+0.56$ & $-0.38$ \\
Zero-Shot Chain-of-Thought & $-2.27$ & $-2.81$ \\
Robustness Vector ($w{=}0.5$) & $-0.16$ & $-0.52$ \\
\bottomrule
\end{tabular}

%% file: tables/tdistraction.tex
\begin{tabular}{l|cc|cc}
\toprule
 & \multicolumn{2}{c|}{Qwen2.5-VL-3B} & \multicolumn{2}{c}{LLaVA-NeXT-8B} \\
Intervention & $\Delta$ V-Distr & $\Delta$ t-Distr & $\Delta$ V-Distr & $\Delta$ t-Distr \\
\midrule
Base Rate & \multicolumn{2}{c|}{0.028 / 0.020} & \multicolumn{2}{c}{0.024 / 0.030} \\
\midrule
Prompt Instruction & $-14\%$ & $+1.0$ & $0\%$ & $-0.1$ \\
Zero-Shot Chain-of-Thought & $+82\%$ & $+15.4$ & $+331\%$ & $+16.4$ \\
M3ID Contrastive Decoding$^{\dagger}$ & $+11\%$ & $+0.6$ & $-22\%$ & $+2.0$ \\
Robustness Vector ($w{=}0.5$) & $\mathbf{-20\%}$ & $\mathbf{+0.3}$ & $\mathbf{-39\%}$ & $\mathbf{+2.4}$ \\
\bottomrule
\end{tabular}

%% file: tables/acc_deltas.tex
\begin{tabular}{l|ccc|ccc|ccc}
\toprule
 & \multicolumn{3}{c|}{\dsbase{} (Test)} & \multicolumn{3}{c|}{\dsname-Human} & \multicolumn{3}{c}{\asmname{} (Held-Out)} \\
Backbone & Base & TV & $\Delta$ (pp) & Base & TV & $\Delta$ (pp) & Base & TV & $\Delta$ (pp) \\
\midrule
Qwen2.5-VL-7B    & 0.90 & 0.93 & $+2.4$ & 0.86 & 0.87 & $+1.0$ & 0.90 & 0.90 & $-0.1$ \\
InternVL3-8B     & 0.93 & 0.94 & $+1.6$ & 0.86 & 0.87 & $+0.6$ & 0.95 & 0.94 & $-0.6$ \\
Qwen2.5-VL-3B    & 0.86 & 0.89 & $+2.7$ & 0.79 & 0.87 & $+8.0$ & 0.89 & 0.89 & $+0.1$ \\
LLaVA-OV-7B      & 0.85 & 0.91 & $+6.4$ & 0.84 & 0.85 & $+0.6$ & 0.91 & 0.91 & $-0.5$ \\
LLaVA-NeXT-8B    & 0.77 & 0.82 & $+5.1$ & 0.72 & 0.77 & $+5.4$ & 0.88 & 0.88 & $-0.4$ \\
Qwen2-VL-2B      & 0.81 & 0.87 & $+5.4$ & 0.74 & 0.82 & $+7.2$ & 0.78 & 0.80 & $+1.4$ \\
LLaVA-1.5-7B     & 0.68 & 0.75 & $+7.2$ & 0.58 & 0.68 & $+10.0$ & 0.73 & 0.77 & $+4.6$ \\
\midrule
\emph{Average}     &  & & $\mathbf{+4.4}$ &  & & $\mathbf{+4.7}$ &  & & $\mathbf{+0.6}$ \\
\bottomrule
\end{tabular}